\documentclass{article}

\usepackage[main, final]{neurips_2026}

\usepackage[utf8]{inputenc} % allow utf-8 input
\usepackage[T1]{fontenc}    % use 8-bit T1 fonts
\usepackage{hyperref}       % hyperlinks
\usepackage{url}            % simple URL typesetting
\usepackage{booktabs}       % professional-quality tables
\usepackage{amsfonts}       % blackboard math symbols
\usepackage{nicefrac}       % compact symbols for 1/2, etc.
\usepackage{microtype}      % microtypography
\usepackage{xcolor}         % colors
\usepackage{graphicx}
\usepackage{wrapfig}
\usepackage{amsmath}
\usepackage{algorithm}
\usepackage{algorithmic}
\usepackage{caption}
\usepackage{enumitem}
\usepackage[normalem]{ulem}

\setcitestyle{round,aysep={,},yysep={;}}

\title{Learning a Unified Cross-Model Semantic Dictionary via Gated Bottleneck Sparse Autoencoders}

\author{%
  Jinchi Zhu\\
  Department of Computer Science\\
  University of Kentucky\\
  Lexington, KY 40506\\
  \texttt{jzh321@uky.edu}
  \And
  Tie Luo\thanks{Corresponding author.}\\
  Department of Electrical and Computer Engineering\\
  \& Department of Computer Science\\
  University of Kentucky\\
  Lexington, KY 40506\\
  \texttt{t.luo@uky.edu}
}

\begin{document}

\maketitle

\begin{abstract}
  Foundation models pretrained on similar data distributions tend to learn overlapping semantic structure, but their representations are distinct due to different model architecture, pretraining objectives and protocols. This prevents unified interpretability and cross-model feature reuse. Existing approaches extract interpretable concepts from individual models and {\em align} these concepts in a shared coordinate system using cross-reconstruction objectives, but the alignment only applies to latent coordinates while the concepts themselves remain model-specific and cannot be shared or reused.
  %\blue{3 sets of concepts each belonging to one model, and try to align the semantic of the same index atom through training. For example, given the same input, USAE/SPARC aims to ensure that the SAEs of three different models activate latents with same indices, thereby aligning concepts across models. But the dictionary atoms of these concepts (i.e. the concept vectors) remain specific to each model and cannot be directly shared across models.} 
  %However, model-specific biases \teal{are not handled specifically and remain entangled with the learned concepts, which limits cross-model reconstruction fidelity, unified interpretability, and feature transferability.} %\blue{“limited cross-model reconstruction fidelity” refers to the capacity for “unified interpretability”: higher reconstruction fidelity indicates stronger unified interpretability. “poor feature transferability” refers to “cross-model feature reuse”, which has two aspects: 1st is as an extension of unified interpretability, it considers whether learned interpretable concepts can generalize(or be reused) to unseen models, enabling broader "unified". 2nd, it examines whether the features reconstructed by these concepts can be used to normal downstream tasks by transfer learning.}
  This paper introduces the {\em Gated Bottleneck Sparse Autoencoder} (GB-SAE), which casts unified cross-model interpretability as a representation factorization problem and solves it by simultaneously learning a unified shared dictionary and multiple private model-specific dictionaries via a gated bottleneck sparse autoencoder framework. The shared dictionary encodes the cross-model common semantic structure in unified concepts that are reusable across all participating models (and beyond), while each private dictionary comprises residual concepts that capture model-specific biases. We evaluate GB-SAE with distinct foundation models including ViT, SigLIP, DINOv2, and ConvNeXt. GB-SAE demonstrates superior reconstruction capability, unified and faithful interpretability, strong reusability and generalization to unseen models and downstream tasks. The code is available at https://github.com/trustAIoT/GB-SAE\_NeurIPS. 
\end{abstract}

\section{Introduction}

Different neural networks trained on sufficiently rich data to perform similar tasks may independently learn representations of the same underlying structure of reality. This is referred to as the Platonic representation hypothesis \citep{huh2024position}, and the condition applies to modern foundation models. Moreover, if their pretraining datasets have similar distributions, this phenomenon becomes more pronounced 
\citep{bommasani2021opportunities, dravid2023rosetta, lu2025representation}.

This suggests that there may exist a common conceptual substrate---composed of human-understandable semantic concepts---that underlies different models despite their heterogeneity. We can therefore exploit this opportunity to create new designs that enable unified model interpretability, feature reuse, and generalization. However, the internal numerical representations of different models are distinct, encoding semantics in their own native ways that are incompatible with each other; each model's representations encode its model-specific biases due to its architecture, pretraining objective, and learning protocol. Consequently, semantic correspondence alone does not warrant successful acquisition of a unified interpretability framework.

Recent research efforts, including USAE~
\citep{thasarathan2025universal} and SPARC~\citep{nasiri-sarvi2026sparc}, approached this problem by learning model-specific sparse concepts and aligning them in a shared latent coordinate system using cross-reconstruction objectives. However, such alignment-based methods only align the latent coordinates/indexes and do not change the nature that semantic concepts are still respectively encoded in each model's own, native representation spaces. As a result, concepts are still entangled with model-specific biases and cannot be shared or reused.

%Fig.~\ref{intro} illustrate this idea conceptually.
% \begin{wrapfigure}{r}{0.5\textwidth}
%   \centering
%   \includegraphics[width=\linewidth]{intro.pdf}
%   \caption{Conceptual illustration of cross-model semantic factorization. Different neural networks encode overlapping semantic concepts entangled with model-specific biases; GB-SAE aims to explicitly recover shared semantic concepts while isolating architecture-dependent variations.}
%   \label{intro}
% \end{wrapfigure}
Our goal is to learn a unified, model-agnostic dictionary of semantic concepts that transcend the idiosyncrasies of individual models. Fulfilling this goal could establish a foundation for unified interpretability, cross-model reuse of semantic concepts, and generalization to unseen models. To this end, we formulate a representation factorization problem and propose a cross-model concept discovery framework called \textbf{Gated Bottleneck Sparse Autoencoder} (GB-SAE). It maps representations from $N$ homogeneous or heterogeneous vision models into a common latent space using a shared bottleneck encoder, and routes the latent activations into a shared branch and a private branch using a learnable gate. The shared branch learns a unified cross-model semantic dictionary while the private branch learns $N$ model-specific residual dictionaries. The competition between the two branches is driven by a shared and a total reconstruction loss, which encourage high-fidelity reconstruction of representations across distinct models; at the same time, a cross-model alignment loss promotes consistency among shared-concept activations of different models in the common latent space. 

We evaluate GB-SAE on ImageNet-1K representations from ViT, SigLIP, and DINOv2, with additional configurations incorporating ConvNeXt. The results show that GB-SAE learns a shared dictionary that provides a unified and model-agnostic conceptual substrate for interpretation, reuse, and generalization: 
\begin{itemize}[leftmargin=2em,itemsep=2pt,topsep=0pt]
    \item {\em Reconstruction power:} Despite distinct and incompatible feature representations encoded by different models, GB-SAE's single dictionary reconstructs all of them accurately (54\% improvement over predicting the average of all the features). %reduction of squared prediction error relative to the mean-prediction baseline
    %achieves a mean shared-branch \(R^2\) of 0.54, explaining approximately 54% of the variance relative to a mean-prediction baseline.
    \item {\em Unified concepts and faithful interpretability:} (1) human-understandable semantic concepts are clearly revealed by visualizing the atoms of the shared dictionary; (2) faithfulness of the shared concepts is evidenced by strong sufficiency and comprehensiveness metrics; (3) unified dictionary is demonstrated by using GB-SAE reconstructed representations to perform zero-shot cross-model feature retrieval and observing dramatically higher recall (up to two orders of magnitude) than baselines without a unified dictionary.
    \item {\em Reusability and generalization:} (1) reusing the shared dictionary (contained in a decoder) on representations encoded by CLIP---unseen during GB-SAE training---achieves superior reconstruction performance (50\% better than predicting average); (2) reusing the representations reconstructed by the shared dictionary on two downstream tasks produces competitive performance: classification (Acc@1/Acc@5=0.86/0.97) and full cross-model feature retrieval (recall of 0.57-0.99 and mAP of 0.48).
\end{itemize}
%at the same time, the reconstruction is not merely numerical but well preserves semantics, evidenced by accurate classification (Acc@1/Acc@5=0.86/0.97) using the reconstructed features;

\section{Related work}
\subsection{Concept-based Interpretability and Sparse Autoencoders}
Concept-based interpretability explains model behavior through human-understandable semantic concepts, typically represented as directions or vector in representation space for attribution or intervention, as in TCAV~\citep{kim2018interpretability}, CBM~\citep{koh2020concept}, and SENN~\citep{alvarez2018towards}. Since these methods often require predefined concepts, subsequent work explores automatic concept discovery from representations \citep{ghorbani2019towards,zhang2021invertible,fel2023craft}, and others build intrinsically interpretable models with learned internal concept variables \citep{espinosa2022concept,schrodi2025selective}.

Sparse Autoencoders (SAEs) offer an unsupervised framework for discovering sparse, often overcomplete, latent features that can be interpreted as semantic concepts \citep{makhzani2013k}. Their sparse decomposition structure has made them useful for analyzing neural representations \citep{ICLR2024_1fa1ab11,rajamanoharan2024jumping}, with extensions such as Matryoshka sparse autoencoders \citep{zaigrajew2025interpreting}, archetypal sparse autoencoders \citep{pmlr-v267-fel25a}, and the analysis of semantic concepts in vision-language models \citep{papadimitriou2025interpreting}. However, these SAE-based methods remain model-specific and do not discover unified or shared cross-model concepts.

\subsection{Cross-Model Concept Alignment}
As multimodal and multi-model systems proliferate, single-model interpretability becomes insufficient in those settings, motivating methods that aim to align semantic concepts across models. Early approaches use post-hoc concept mining and matching but face scalability and flexibility limitations \citep{dravid2023rosetta,kowal2024understanding}.

Recent SAE-based methods have advanced cross-model concept discovery by mapping activations from different architectures into a shared sparse latent space. Universal Sparse Autoencoders (USAE) \citep{thasarathan2025universal}  trains model-specific encoder-decoder pairs by randomly selecting one model's encoder per iteration to map its activations into the shared latent space, which is then decoded across all model decoders, to achieve soft conceptual alignment (in terms of coordinates only). SPARC \citep{nasiri-sarvi2026sparc} strengthens this alignment by introducing a global topK activation mechanism to enforce identical latent indices across streams to enhance semantic consistency.
%USAE aligns model-specific SAEs by sharing sparse activations during cross-reconstruction \citep{thasarathan2025universal}, while SPARC adds explicit constraints to enforce semantic consistency between learned dictionaries \citep{nasiri-sarvi2026sparc}. These methods improve scalability and establish effective cross-model concept correspondence.

However, alignment-based approaches at best learn a shared coordinate system that aligns concepts by indexes only, while concepts themselves are still maintained separately by individual models, limiting unified interpretability and representation reuse; a separate encoder-decoder pair has to be maintained for each specific model.

\subsection{Representation Alignment and Unified Representation}
Representation alignment maps representations from different models or modalities into a shared embedding space. It is exemplified by CLIP \citep{radford2021learning} and has been widely used across tasks \citep{luo2020univl,ramesh2021zero,cho2021unifying,artetxe2020cross}. 

Such alignment is possible because large-scale pretrained models tend to encode shared semantic structure, although in model-specific forms \citep{ICLR2024_44a2b9f7}. This phenomenon has been interpreted through the Platonic representation hypothesis, while such underlying capacity of foundation models is referred to as \emph{representation potential} \citep{lu2025representation}. As more models are aligned, the shared space can move toward a more unified representation \citep{huh2024position}. However, the key question of {\em how} to achieve this remains unanswered. Furthermore, existing approaches \citep{Girdhar_2023_CVPR} that aim to learn unified representations predominantly produce dense representations optimized for downstream or task-specific performance rather than interpretability. It thus remains largely unexplored for developing an interpretability framework with a unified concept dictionary.

\section{Method}

\subsection{Objective Definition}\label{sec:objective definition}
Let $\mathcal{F} = \{f_1,\ldots,f_N\}$ be a collection of pretrained neural networks and let $\mathcal{X}$ denote a dataset. For each $f_i \in \mathcal{F}$ and input $x \in \mathcal{X}$, we extract an internal representation
$\mathbf{a}_i = f_i(x) \in \mathbb{R}^{n \times d_i}$,
where $n$ denotes the number of activation instances (e.g., tokens) and $d_i$ is the feature dimensionality.

Our objective is to learn: (i) a \emph{shared} concept dictionary $\mathbf{D}^{\mathrm{sh}} \in \mathbb{R}^{h \times d'}$, and (ii) a set of \emph{private} dictionaries $\mathbf{D}^{\mathrm{pr}}_i \in \mathbb{R}^{h \times d'}, i=1,\ldots,N$, where $d'$ is the common dimension projected from all the $\mathbf{a}_i$'s, and $h$ is the latent dimension after passing the representations to a shared encoder. Each atom in $\mathbf{D}^{\mathrm{sh}}$ is intended to capture a semantic concept that is consistently expressed across participating models, providing a shared latent semantic vector that can be consistently interpreted and reused across these models and potentially generalized to unseen models. In contrast, each \(\mathbf D_i^{\mathrm{pr}}\) captures residual semantic concepts not represented by the shared dictionary, with its atoms encoding model-specific variations arising from architectural choices, training objectives, or learned parameters.

Formally, we seek to factorize each $\mathbf{a}_i$ into a shared and a private component:
\[ \mathbf{a}'_i \approx \mathbf{z}_i^{\mathrm{sh}} \mathbf{D}^{\mathrm{sh}} + \mathbf{z}_i^{\mathrm{pr}} \mathbf{D}^{\mathrm{pr}}_i, \]
where $a'_i\in\mathbb R^{n\times d'}$ is $a_i$ after a fixed projection described in Section \ref{sec:proj}, and $\mathbf{z}_i^{\mathrm{sh}}, \mathbf{z}_i^{\mathrm{pr}} \in \mathbb{R}^{n \times h}$ are sparse activation coefficients.

\begin{figure}
  \centering
  \includegraphics[width=1.0\linewidth]{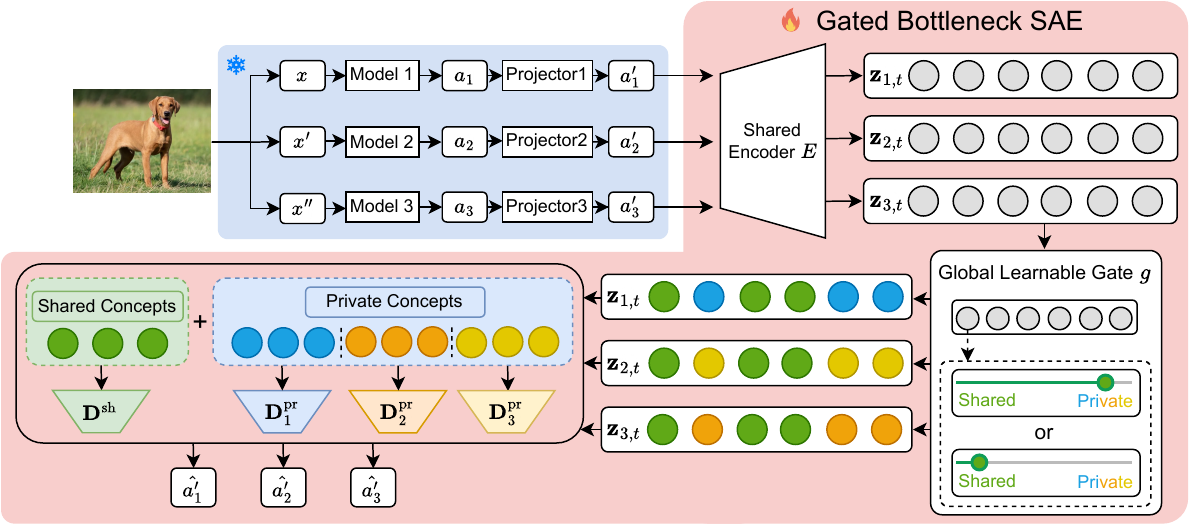}
  \caption{Overall architecture of GB-SAE. Features from multiple frozen backbone models are projected into a common space and encoded by a shared sparse encoder. A global learnable gate partitions latent dimensions into shared and private branches, yielding a model-agnostic semantic dictionary and model-specific residual dictionaries. The illustrated routing is shown for a single token; in practice, the gate is a global vector broadcast across tokens.}
  \label{pipeline}
\end{figure}

\subsection{GB-SAE Architecture}
\subsubsection{Feature Extraction and Projection}\label{sec:proj}
As shown in Fig.~\ref{pipeline}, multiple pretrained backbones $f_i$ extract semantic representations $\mathbf{a}_i = f_i(x) \in \mathbb{R}^{n \times d_i}$ from an input sample $x$. Since feature dimensionalities $d_i$ differ across models, we introduce a {\em frozen} random linear projector to map each representation into a common space:
\[
\mathbf{a}'_i = \mathrm{Projector}_i(\mathbf{a}_i), \quad \mathrm{Projector}_i:\mathbb{R}^{d_i} \rightarrow \mathbb{R}^{d'}.
\]
The fixed random projection maps each model's representations to a common dimensionality while approximately preserving their within-model pairwise geometry under the Johnson–Lindenstrauss lemma. It does not itself align representations across different models; this alignment is subsequently learned through the shared encoder and cross-model alignment objective.
%Thus, we reformulate the factorization objective in Section~\ref{sec:objective definition} as $\mathbf{a}'_i \approx \mathbf{z}_i^{\mathrm{sh}} \mathbf{D}^{\mathrm{sh}} + \mathbf{z}_i^{\mathrm{pr}} \mathbf{D}^{\mathrm{pr}}_i$.
Next, the projected features $\mathbf{a}'_i \in \mathbb{R}^{n \times d'}$ are processed by a shared encoder. 

\subsubsection{Shared Encoder and Learnable Gate}
% To obtain sparse latent codes in a common semantic space, we apply a shared encoder $E_{\theta}$:
% \[
% \mathbf{z}_i = E_{\theta}(\mathbf{a}'_i) 
% = \mathrm{TopK}(W_{\mathrm{enc}}\mathbf{a}'_i - b_{\mathrm{enc}}),
% \quad \mathbf{z}_i \in \mathbb{R}^{n \times h}, \quad h>>d'.
% \]
% In this process, the Top-K operation selects dimensions with the highest activation values among $h$ dimensions and sets the activations of all other dimensions to zero, thereby producing sparse latent codes.
To map representations from different models into a common latent space, we employ a shared encoder $E_\phi$ as a bottleneck that transforms heterogeneous representations into sparse latent codes. Letting $\mathbf{a}'_{i,t} := (\mathbf{a}'_i)_{t,:}^{\top}\in \mathbb{R}^{d'}$ denote the $t$-th token feature from model $i$, we compute
\[
% \mathbf{z}_{i,t}
% = E_\theta(\mathbf{a}'_{i,t})
%  = \operatorname{TopK}\!\left(
%      \mathbf{W}_{\mathrm{enc}}\mathbf{a}'_{i,t}
%      + \mathbf{b}_{\mathrm{enc}}
%    \right),
\begin{aligned}
\mathbf{z}_{i,t}
&=E_\phi(\mathbf{a}'_{i,t})
=\operatorname{TopK}\!\left(
    \operatorname{ReLU}(
        \mathbf{W}_{\mathrm{enc}}\mathbf{a}'_{i,t}
        + \mathbf{b}_{\mathrm{enc}})
                        \right), \\
&t=1,\ldots,n,
\quad \mathbf{z}_i
= [\mathbf{z}_{i,1}, \ldots, \mathbf{z}_{i,n}]^{\top}
% \begin{bmatrix}
% \mathbf{z}_{i,1}^{\top}\\
% \vdots\\
% \mathbf{z}_{i,n}^{\top}
% \end{bmatrix}
\in \mathbb{R}^{n\times h},
\end{aligned}
\]
where $\mathbf{W}_{\mathrm{enc}}\in\mathbb{R}^{h\times d'}$ and $\mathbf{b}_{\mathrm{enc}}\in\mathbb{R}^{h}$, with $h\gg d'$. For each token, $\operatorname{TopK}$ retains the $K$ largest encoder activations among the $h$ latent dimensions and sets all others to zero. The same encoder parameters are used for all tokens and models, and the resulting token-level codes form the rows of $\mathbf z_i$. While $\mathbf{z}_i$ resides in a unified latent space, it still entangles shared and model-specific semantics. We therefore introduce a global learnable gate $g$, which softly partitions the $h$ latent dimensions:
\[
\mathbf{g} = \sigma(\theta) \in (0,1)^h
\; \rightarrow\; 
\mathbf{z}^{\mathrm{sh}}_i = \mathbf{z}_i \odot \mathbf{g}, \;
\mathbf{z}^{\mathrm{pr}}_i = \mathbf{z}_i \odot (1-\mathbf{g}).
\]
Dimensions with $\mathbf{g}_j \approx 1$ are routed to the shared branch, while those with $\mathbf{g}_j \approx 0$ are routed to the private branch. To encourage effective separation, we impose the following regularization term:
\[
\mathcal{L}_{\mathrm{gate}}= \tfrac{1}{h}\sum_{j=1}^{h} \mathbf{g}_j(1-\mathbf{g}_j).
\]

\subsubsection{Multi-Decoder Reconstruction}
GB-SAE employs one shared {\em decoder} $\mathbf{D}^{\mathrm{sh}}$ and $N$ private decoders $\mathbf{D}^{\mathrm{pr}}_i$ to jointly reconstruct all the projected representations $\mathbf{a}'_i$, using the sparse latent codes $\mathbf{z}_i$.
Specifically, $\mathbf{D}^{\mathrm{sh}}$ reconstructs shared representations from $\mathbf{z}^{\mathrm{sh}}_i$ in a cross-model shared representation space, while each $\mathbf{D}^{\mathrm{pr}}_i$ reconstructs from $\mathbf{z}^{\mathrm{pr}}_i$ private representations in a model-specific private representation space. By this design,  $\mathbf{D}^{\mathrm{sh}}$'s atoms are optimized to capture model-agnostic semantic structure, i.e., shared concepts, while each $\mathbf{D}^{\mathrm{pr}}_i$ captures model-specific concepts that reflect biases not encoded by $\mathbf{D}^{\mathrm{sh}}$. The overall reconstruction process can be formulated as:
\[
\hat{\mathbf{a}}'_i
= \mathbf{z}^{\mathrm{sh}}_i \mathbf{D}^{\mathrm{sh}}
+ \mathbf{z}^{\mathrm{pr}}_i \mathbf{D}^{\mathrm{pr}}_i,
\qquad
\hat{\mathbf{a}}'_i \in \mathbb{R}^{n \times d'}.
\]

\subsection{Training Objective}\label{sec:training_objective}
Training GB-SAE aims to achieve three goals simultaneously: (i) accurate reconstruction of the projected features, (ii) prevention of collapse where private branch dominates the factorization and absorbs all information, and (iii) consistent alignment of shared semantics across participating models.

\paragraph{Reconstruction.}
We define two squared Frobenius reconstruction losses:
\begin{align*}
\mathcal{L}_{\mathrm{recon}}^{\mathrm{total}}
= \frac{1}{n d'}\sum_{i=1}^{N} \left\| \mathbf{a}'_i - \hat{\mathbf{a}}'_i \right\|_F^2,
\qquad
\mathcal{L}_{\mathrm{recon}}^{\mathrm{sh\text{-}only}}
= \frac{1}{n d'}\sum_{i=1}^{N} \left\| \mathbf{a}'_i - \mathbf{z}^{\mathrm{sh}}_i \mathbf{D}^{\mathrm{sh}} \right\|_F^2.
\end{align*}
The second loss encourages the shared branch to be predictive while discouraging the private decoders from dominating reconstruction.

\paragraph{Cross-model Alignment.}
To enforce semantic consistency of the shared codes, we align the sparse activations across all model pairs. For any model pair $(i,j)\in\mathcal{P}$ and token $t\in\{1,\ldots,n\}$, let $\mathbf{z}^{\mathrm{sh}}_{i,t}\in\mathbb{R}_{\ge 0}^{h}$ and $\mathbf{z}^{\mathrm{sh}}_{j,t}\in\mathbb{R}_{\ge 0}^{h}$ denote the shared latent activations. We first normalize the shared activations within each token as $\bar{\mathbf{z}}^{\mathrm{sh}}_{i,t}$ and $\bar{\mathbf{z}}^{\mathrm{sh}}_{j,t}$, and then define a continuous Jaccard similarity as follows:
\[
\operatorname{IoU}
\left(
\mathbf{z}^{\mathrm{sh}}_{i,t},
\mathbf{z}^{\mathrm{sh}}_{j,t}
\right)
=
\frac{
\sum_{k=1}^{h}
\min
\left(
\bar{\mathbf{z}}^{\mathrm{sh}}_{i,t}(k),
\bar{\mathbf{z}}^{\mathrm{sh}}_{j,t}(k)
\right)
}{
\sum_{k=1}^{h}
\max
\left(
\bar{\mathbf{z}}^{\mathrm{sh}}_{i,t}(k),
\bar{\mathbf{z}}^{\mathrm{sh}}_{j,t}(k)
\right)
+\epsilon
},
\]
where $k$ denotes the concept latent index.
To further align semantic intensity on the shared support, we introduce an overlap-weighted magnitude discrepancy. The overlap weight for latent $k$ is defined as
\[
w_{ij,t}(k)
=
\operatorname{sg}
\left[
\min
\left(
\bar{\mathbf{z}}^{\mathrm{sh}}_{i,t}(k),
\bar{\mathbf{z}}^{\mathrm{sh}}_{j,t}(k)
\right)
\right],
\]
where $\operatorname{sg}[\cdot]$ denotes the stop-gradient operator. The magnitude discrepancy is then
\[
\Delta_{\mathrm{mag}}
\left(
\mathbf{z}^{\mathrm{sh}}_{i,t},
\mathbf{z}^{\mathrm{sh}}_{j,t}
\right)
=
\frac{
\sum_{k=1}^{h}
w_{ij,t}(k)
\left|
\mathbf{z}^{\mathrm{sh}}_{i,t}(k)
-
\mathbf{z}^{\mathrm{sh}}_{j,t}(k)
\right|
}{
\sum_{k=1}^{h} w_{ij,t}(k)+\epsilon
}.
\]
The cross-model alignment loss is defined as
\[
\mathcal{L}_{\mathrm{align}}
=
\frac{1}{|\mathcal{P}|}
\sum_{(i,j)\in\mathcal{P}}
\mathbb{E}_{t=1..n}
\left[
1-
\operatorname{IoU}
\left(
\mathbf{z}^{\mathrm{sh}}_{i,t},
\mathbf{z}^{\mathrm{sh}}_{j,t}
\right)
+
\lambda_{\mathrm{mag}}
\Delta_{\mathrm{mag}}
\left(
\mathbf{z}^{\mathrm{sh}}_{i,t},
\mathbf{z}^{\mathrm{sh}}_{j,t}
\right)
\right],
\]
where $\lambda_{\mathrm{mag}}$ controls the strength of magnitude alignment. This objective jointly encourages the shared latents to activate on consistent sparse supports across different models and also to activate with consistent magnitudes on the overlapping shared support.
%just MSE may work, or bound it as $\exp(-\frac{(z_i - z_j)^2}{2\sigma^2}) \in(0,1]$ where $\sigma$ controls how quickly the score decreases when the two activations differ. We can call this core element-wise agreement score.

\paragraph{Overall Objective.}
The full training loss is
\[
\begin{aligned}
\mathcal{L}_{\mathrm{total}}
= \;\mathcal{L}_{\mathrm{recon}}^{\mathrm{total}} + \lambda_{\mathrm{sh\text{-}only}}\mathcal{L}_{\mathrm{recon}}^{\mathrm{sh\text{-}only}}
+ \lambda_{\mathrm{align}}\mathcal{L}_{\mathrm{align}} + \lambda_{\mathrm{gate}}\mathcal{L}_{\mathrm{gate}}.
\end{aligned}
\]

\section{Experiments}
We evaluate GB-SAE in the visual domain using ImageNet-1K and three foundation-model configurations:
\begin{itemize}[leftmargin=2em,itemsep=1pt,topsep=0pt]
    \item Configuration I: ViT \citep{dosovitskiy2021an}, SigLIP \citep{zhai2023sigmoid}, and DINOv2 \citep{oquab2024dinov}. These models are pretrained with substantially different protocols and objectives.
    \item Configuration II: ViT, ConvNeXt \citep{liu2022convnet}, and DINOv2. This setting includes a CNN backbone to introduce inductive biases fundamentally different from transformers.
    \item Configuration III: consists of all the four backbones—ViT, SigLIP, DINOv2, and ConvNeXt—to evaluate GB-SAE with a more comprehensive ensemble.
\end{itemize}
For each backbone, we extract the final-layer feature representations and feed them directly into GB-SAE without any alteration. Additional implementation details are provided in Appendix~\ref{app:impl}. We report the results for Configuration I in the main paper and provide the results for Configurations II and III in Appendix~\ref{app:configs}.

\subsection{Reconstruction Capability}\label{sec:reconstruct}
The first important and direct measure of GB-SAE's factorization is the fidelity in reconstructing each model's representations using GB-SAE's learned dictionaries, especially the shared one. The metric we use is the coefficient of determination ($R^2$), defined as $R^2(\mathbf{a},\hat{\mathbf{a}}) = 1 - \frac{\|\mathbf{a} - \hat{\mathbf{a}}\|_F^2}{\|\mathbf{a} - \bar{\mathbf{a}}\|_F^2}\in(-\infty,1]$, where $\hat{\mathbf{a}}$ denotes the reconstructed $\mathbf{a}$ and $\bar{\mathbf{a}}$ is the the mean of all the $\mathbf{a}$'s under testing. The reconstruction target $\mathbf{a}$ can be (1) an original representation $\mathbf{a}_i$, (2) a projected representation $\mathbf{a}'_i$, or (3) a residual representation $\mathbf{a} - \mathbf{z}^{\mathrm{sh}}_i \mathbf{D}^{\mathrm{sh}}$ where $\mathbf{a}$ is either $\mathbf{a}_i$ or $\mathbf{a}'_i$. Higher $R^2$ values indicate better reconstruction (i.e., preserving more complete semantic information). When calculating $R^2$ for the raw $\mathbf{a}_i$, an inverse projection is applied to $\hat{\mathbf{a}}$ to map it to the same dimension as $\mathbf{a}_i$.

In Table~\ref{tab:recon}(a), we report $R^2$ for reconstruction using the shared dictionary only ($R^2$-sh), the private dictionary only ($R^2$-pr), and both ($R^2$-total); we also reconstruct residuals using the private branch and report $R^2$-res. The results show that GB-SAE achieves cross-model $R^2$-total $=0.60$ and $R^2$-sh $=0.54$ (average of projected and raw; the same applies below too). The closeness between the two values indicates that the shared dictionary alone retains most of the measured reconstruction performance, %the shared dictionary accounts for the majority ($\sim 90\%$) of the overall reconstruction performance, 
and also hints toward a strong cross-model reusability (further evaluated in Section \ref{sec:reuse}). An $R^2$-total of 0.60 means that our reconstruction improves 60\% over predicting the mean of all representations. The private dictionaries have their mean $R^2$-pr $=0.11$, indicating that they contribute much less to the reconstruction than the shared branch. On the other hand, the private branch fares better when reconstructing residuals (mean $R^2$-res $=0.135$), reflecting its specialization in capturing model-specific biases. 
Moreover, adding the private branch improves reconstruction over the shared branch alone, consistent with its intended role of complementing the shared reconstruction.

Table~\ref{tab:recon}(b) reports baseline methods' performance. Because prior methods perform {\em pair-wise} reconstruction, without a shared and reusable component like GB-SAE, so direct comparison on exactly the same metrics as in Table~\ref{tab:recon}(a) is not possible. We address this by defining the closest corresponding metrics for the baselines: (1) $R^2$-self measures how well a model reconstructs its own representations using its own learned dictionary (i.e., $\mathbf{a}_i$ vs. $\hat{\mathbf{a}_i}=\mathbf{z}_i \mathbf{D}_i$), which corresponds to (and is presumably higher than) GB-SAE's $R^2$-total; in this case, a single-model standard SAE ($R^2$-self$=0.71$) serves as an empirical upper bound; (2) $R^2$-cross evaluates cross-model reconstruction (i.e., $\mathbf{a}_i$ vs. $\mathbf{z}_j \mathbf{D}_i$ for $i \neq j$), which corresponds to GB-SAE's $R^2$-sh. The table shows that GB-SAE's $R^2$-total is comparable to USAE and SPARC self-reconstruction, while having the $R^2$-sh substantially better than baselines' cross-reconstruction, which (the latter) matters the most to unified interpretability.

\begin{table}[t]
  \caption{\small (a) Reconstruction performance of GB-SAE on the ImageNet validation set, reported for both projected and raw representations (projected / raw). (b) Reconstruction performance of baselines, reported as mean token-level $R^2$ (same as in table (a)), where $R^2$-self and $R^2$-cross approximately correspond to GB-SAE's $R^2$-total and $R^2$-sh, respectively.}
  \label{tab:recon}
  \centering
  \begin{small}
    \renewcommand{\arraystretch}{0.8}

    \begin{minipage}[t]{0.58\linewidth}
      \centering
      \textbf{(a) GB-SAE reconstruction}
      \vspace{0.35em}

      \resizebox{\linewidth}{!}{%
      \begin{sc}
      \begin{tabular}{lcccc}
        \toprule
        Model & $R^2$-total $\uparrow$ & $R^2$-sh $\uparrow$ & $R^2$-pr $\downarrow$ & $R^2$-res $\uparrow$ \\
        \midrule
        ViT    & 0.56/0.56 & 0.52/0.52 & 0.07/0.07 & 0.09/0.09 \\
        SigLIP & 0.58/0.58 & 0.52/0.52 & 0.12/0.12 & 0.12/0.12 \\
        DinoV2 & 0.65/0.65 & 0.53/0.59 & 0.11/0.16 & 0.24/0.15 \\
        \midrule
        Mean   & 0.60/0.60 & 0.53/0.55 & 0.10/0.12 & 0.15/0.12 \\
        \bottomrule
      \end{tabular}
      \end{sc}
      }
    \end{minipage}
    \hfill
    \begin{minipage}[t]{0.41\linewidth}
      \centering
      \textbf{(b) Self/cross reconstruction}
      \vspace{0.35em}

      \resizebox{\linewidth}{!}{%
      \begin{sc}
      \begin{tabular}{lcc}
        \toprule
        Method & $R^2$-self & $R^2$-cross (pair-wise) \\
        \midrule
        SAE (upp. bnd) & 0.71 & -- \\
        USAE         & 0.61 & 0.20 \\
        SPARC        & 0.55 & 0.41 \\
        \bottomrule
      \end{tabular}
      \end{sc}
      }
    \end{minipage}

  \end{small}
\end{table}

\subsection{Unified Concepts and Faithful Interpretability}

\subsubsection{Visualization of Human-Interpretable Concepts}
Visualization directly assess whether a learned dictionary capture human-interpretable semantics. We design a visualization pipeline for both shared and private GB-SAE dictionary atoms, as described in Appendix~\ref{app:vis} Algorithm~\ref{alg:gbsae_vis}. 

\begin{figure}
  \centering
  \includegraphics[width=\linewidth]{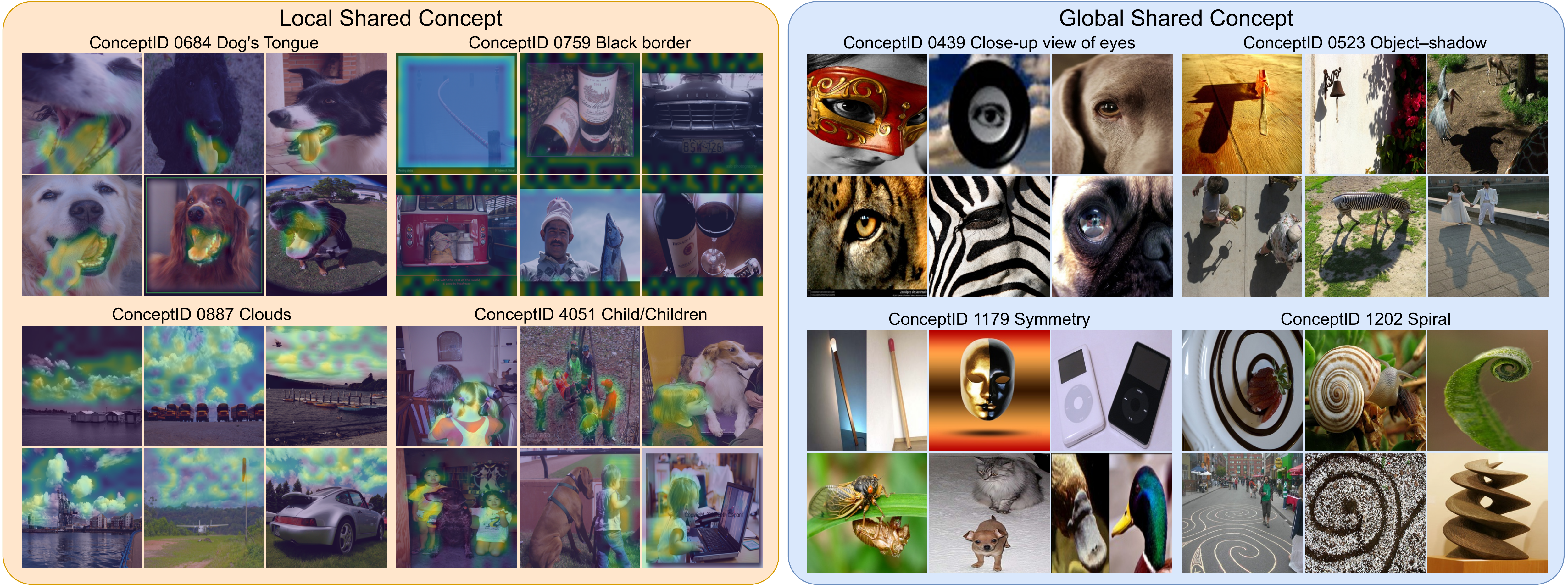}
  \caption{Visualization of shared semantic concepts discovered by GB-SAE across pretrained foundation models. The left panel shows four local concepts, each depicted by six semantically consistent regions across six different images. The right panel shows four global concepts, each depicted within six images; heatmaps are omitted as global concepts are activated throughout nearly the whole image.}
  \label{visual-shared}
\end{figure}

Figure~\ref{visual-shared} visualizes shared concepts discovered by GB-SAE across all participating pretrained models. It shows that the concepts are both semantically meaningful and human-recognizable. For more clarity, we group the concepts into two categories: (1) A \textbf{local shared concept} activates on a {\em local region} that is semantically consistent across images; such concepts capture object-centric or part-based semantics with strong spatial locality; for instance, clouds, children, or distinctive textures. (2) A \textbf{global shared concept} captures holistic structural patterns spanning the entire image; these concepts typically correspond to geometric configurations or relational layouts, reflecting abstract regularities. 

For completeness, we also provide the visualization of private concepts in Appendix~\ref{app:vis} Figure~\ref{fig:visual-private}. In contrast, private atoms exhibit strong dependence on specific models.

\subsubsection{Faithfulness of Interpretability via Concept Attribution Analysis}
We conduct a concept attribution analysis using two faithfulness metrics---Sufficiency and Comprehensiveness---proposed in \citep{10.1145/3711896.3737414}. We design a linear probing experiment as follows. For each model, we train a simple linear classifier---only a single fully-connected layer---on the projected representations for ImageNet1K. For each input image $x$ with ground-truth class $y$, the corresponding logit is $\ell_y(x) = W^\top \bar{\mathbf{a}}'_i+b$, where $W$ denotes the classifier weights and $\bar{\mathbf{a}}'_i$ is mean-pooled $\mathbf{a}'_i$. We then freeze the trained classifier and replace $\bar{\mathbf{a}}'_i$ projected) with its GB-SAE reconstruction $\bar{\hat{\mathbf{a}}}'_i$. Next, we rank the latent activation's dimensions by their contribution to $\ell_y(x)$, and intervene by keeping or removing the top-k dimensions. Accordingly, the Sufficiency and Comprehensiveness are defined as:
\begin{align*}
\mathrm{Suff@}k
= \mathbb{E}\!\left[
\frac{\ell_y^{\mathrm{full}} - \ell_{y,k}^{\mathrm{keep}}}
{\left|\ell_y^{\mathrm{full}}\right|}
\right], \qquad
\mathrm{Comp@}k
= \mathbb{E}\!\left[
\frac{\ell_y^{\mathrm{full}} - \ell_{y,k}^{\mathrm{rem}}}
{\left|\ell_y^{\mathrm{full}}\right|}
\right].
\end{align*}
Smaller Suff@k indicates greater sufficiency of the selected latents, while larger Comp@k reflects stronger reliance on them.%measured on all samples; not better on onlly correct predictions

\begin{figure}
  \centering
  \includegraphics[width=\linewidth]{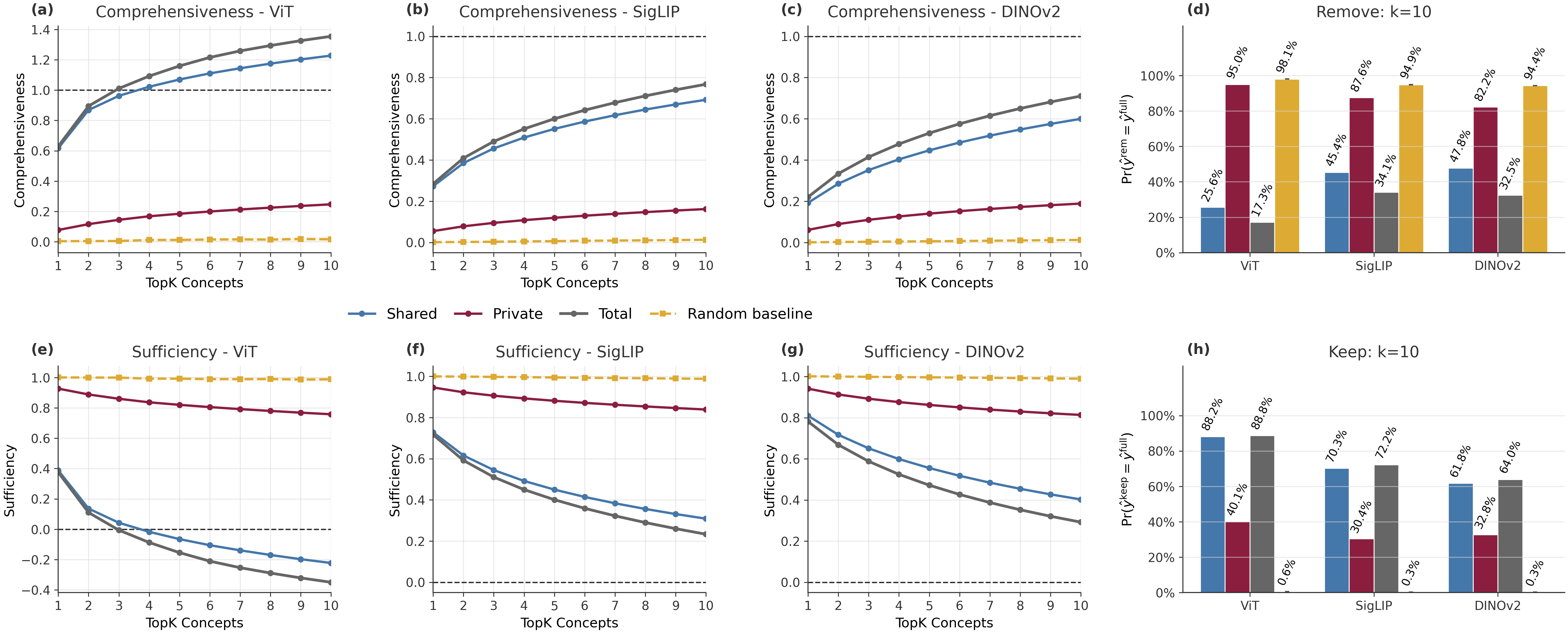}
  \caption{Comprehensiveness (higher is better) and Sufficiency (lower is better) as a function of the number of top-ranked concepts $k$. The bar charts report prediction agreement at $k=10$ on their predicted class $\hat y$: $\Pr(\hat{y}^{\mathrm{rem}}=\hat{y}^{\mathrm{full}})$ after removal (lower is better) and $\Pr(\hat{y}^{\mathrm{keep}}=\hat{y}^{\mathrm{full}})$ after retention (higher is better).}
  \label{fig:suff-comp}
\end{figure}

Figure~\ref{fig:suff-comp} shows that shared concepts yield substantially stronger interpretive power than private concepts, while being close to total concepts, across the evaluated backbones. This is consistent with the intended factorization. All three outperform the random baseline, which uses a dictionary whose atoms are random and orthogonal with each other. This demonstrates that the learned concepts faithfully explain predictive behavior. As the number of kept or removed concepts increases, sufficiency decreases while comprehensiveness increases. Notably, some values extend beyond the conventional [0, 1] range. This is possible because these metrics measure normalized changes in logits rather than probabilities: keeping a subset of concepts can produce a higher target-class logit than the full reconstruction, while removing concepts can result in a logit change exceeding the magnitude of the original full logit.

The bar charts further validate the faithfulness by examining the final predicted classes, rather than examining one particular logit. We see that prediction using ten shared concepts agrees with prediction using the full dictionary in 62–88\% of the time (bottom chart), yet this agreement drops to only 26–48\% upon removing these concepts (upper chart).

Together, these results demonstrate that GB-SAE discovers compact, discriminative concepts that faithfully explain model behavior.

\subsubsection{Unified Dictionary via Zero-shot Cross-model Retrieval}\label{sec:retrieval}
Our learned shared dictionary $\mathbf{D}^{\mathrm{sh}}$ is supposed to define a unified semantic-atom repository across models. This experiment verifies whether $\mathbf{D}^{\mathrm{sh}}$ induces a truly model-agnostic semantic space. To this end, we perform cross-model feature retrieval using representations reconstructed by $\mathbf{D}^{\mathrm{sh}}$, with no additional training. Specifically, given a feature $\mathbf{a}_i$ extracted from backbone $f_i$, we project and encode it using the frozen GB-SAE, and take its reconstructed, shared component only, i.e., $\mathbf{z}^{\mathrm{sh}}_i \mathbf{D}^{\mathrm{sh}}$. Retrieval is then performed in this shared space among all models when $i\neq j$, using cosine similarity (with mean pooling applied beforehand).

For comparison, we establish three baselines. The first is random retrieval. The second performs retrieval using reconstructed private components, i.e., $\mathbf{z}^{\mathrm{pr}}_i \mathbf{D}_i^{\mathrm{pr}}$. The last retrieves directly in the projected feature space, i.e., using $\mathbf{a}_i'$. Since the projector is linear, random, and frozen, its output $\mathbf{a}_i'$ can be viewed as a geometry-preserving surrogate of the original feature $\mathbf{a}_i$.

\begin{figure}
  \centering
  \begin{minipage}[t]{0.49\linewidth}
    \centering
    \includegraphics[height=0.23\textheight,width=\linewidth,keepaspectratio]{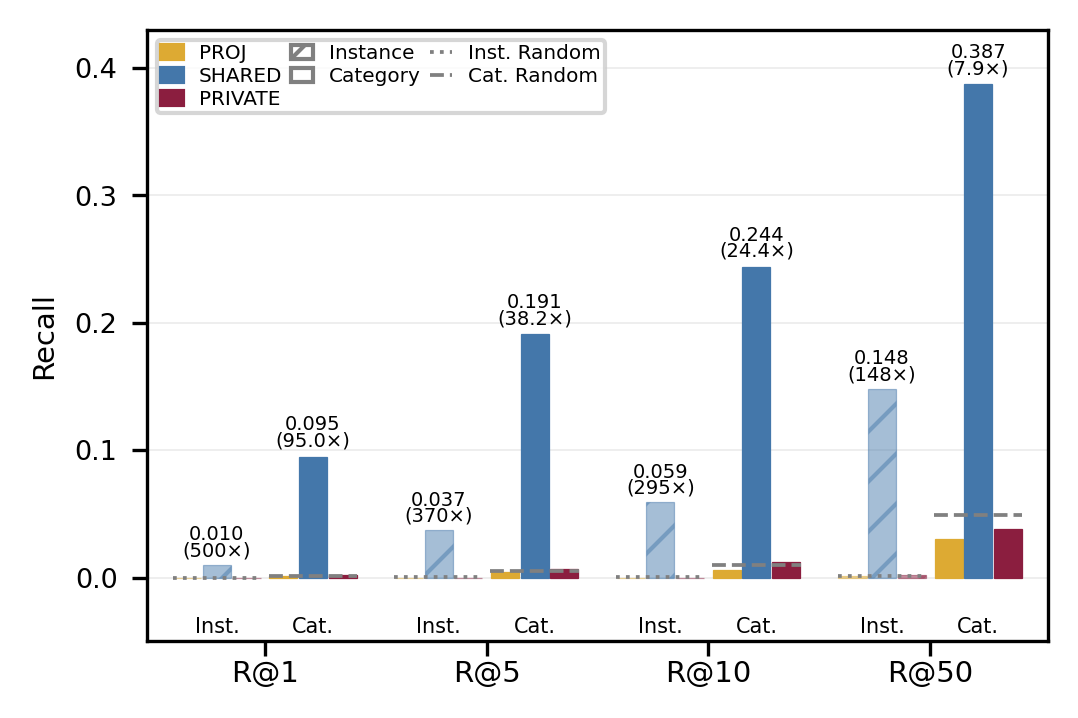}
    \vspace{-15pt}
    \caption{Zero-shot cross-model feature retrieval using $\mathbf{z}^{\mathrm{sh}}_i \mathbf{D}^{\mathrm{sh}}$ (``shared''; light and dark blue bars). We report Recall@$k$ at instance and category levels. Multiples in parentheses denote improvements over random retrieval.}
    \label{fig:retrieval}
  \end{minipage}
  \hfill
  \begin{minipage}[t]{0.47\linewidth}
    \centering
    \includegraphics[height=0.23\textheight,width=\linewidth,keepaspectratio]{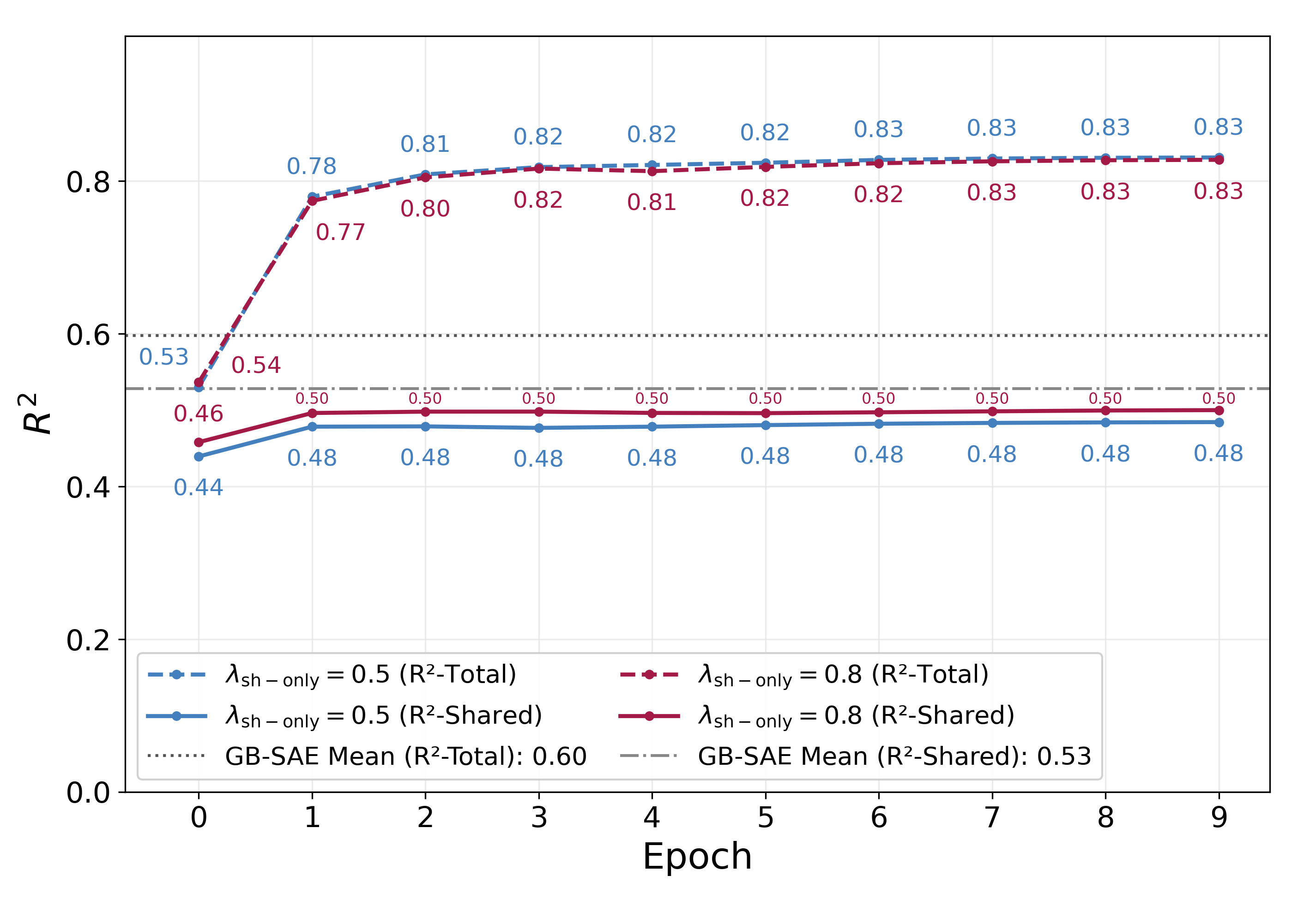}
    \vspace{-15pt}
    \caption{Training curves for generalizing $\mathbf{D}^{\mathrm{sh}}$ to an unseen backbone (CLIP). All components are frozen except for a linear input projector and a CLIP-specific private decoder.}
    \label{clip}
  \end{minipage}
  \vspace{-10pt}
\end{figure}

We consider both instance retrieval and category retrieval, where the former means a retrieval is considered correct only if the instance (image) matches, and the latter means correct as long as the class matches.

Fig.~\ref{fig:retrieval} shows that retrieval using GB-SAE shared representations (blue) dramatically outperforms all baselines, achieving improvement of up to two orders of magnitude. Retrieval using private (red) or projected (orange) features perform nearly the same as random retrieval (gray lines). The absolute values are generally low because of the large number of instances (50,000 validation images) and categories (1,000 classes).

These results suggest that, thanks to the unification and alignment process in GB-SAE, the dictionary $\mathbf{D}^{\mathrm{sh}}$ concentrates most of the transferable semantics, and thereby induces a genuinely aligned semantic space that enables effective cross-model matching without explicit supervision.

\subsection{Reusability and Generalization}\label{sec:reuse}

This section evaluates the reusability and generalization of the unified dictionary and its reconstructed representations.

\subsubsection{Reusing Dictionary and Generalizing to Unseen Models}

This experiment evaluates whether the shared decoder, with dictionary $\mathbf{D}^{\mathrm{sh}}$ being its weights, can be reused as a decoder for other unseen models. Our hypothesis is that, if $\mathbf{D}^{\mathrm{sh}}$ captures a model-agnostic semantic basis, then representations from unseen models should be mappable into this space with minimal, linear adaptation.

We take CLIP, which is unseen during GB-SAE training, and uses its image encoder to obtain image representations to serve as the input $\mathbf{a}_i$ for the pretrained GB-SAE. In GB-SAE, we freeze the encoder, the gating module, and the shared decoder $\mathbf{D}^{\mathrm{sh}}$; naturally, a CLIP-specific private decoder $\mathbf{D}^{\mathrm{pr}}_{\mathrm{CLIP}}$ needs to be added to capture model-specific residual information. The adaptation is replacing the frozen random projector with a learnable linear single-layer adapter, to align CLIP features with the common semantic space.

Our experiment (Figure~\ref{clip}) finds that GB-SAE with CLIP converges quickly within 4--5 epochs, confirming that the adaptation is light, and achieves $\mathrm{R}^2$-total $=0.83/0.83$ and $\mathrm{R}^2$-sh $=0.48/0.50$ (corresponding to $\lambda_{\mathrm{sh\text{-}only}}=0.5/0.8$). This solid shared-branch reconstruction confirms that the representation space induced by $\mathbf{D}^{\mathrm{sh}}$ can effectively express CLIP semantics even though CLIP is not part of the training configuration of GB-SAE, demonstrating that $\mathbf{D}^{\mathrm{sh}}$ can serve effectively as a reusable semantic substrate for models beyond the training ensemble.

\subsubsection{Reusing Features and Generalizing to Downstream Tasks}

To further evaluate the reusability and downstream utility of GB-SAE, we test two downstream tasks by reusing the feature representations from GB-SAE's unified dictionary.

\textbf{Downstream classification.}
We train multiple ImageNet-1K classification heads on features from different sources and compare their performance. These sources are: (i) features output by the frozen projector, $\mathbf{a}'$, regarded as original features; (ii) features reconstructed by prior methods (USAE and SPARC); and (iii) features reconstructed by the shared branch of GB-SAE, i.e., $\bar{\mathbf{z}}^{\mathrm{sh}} \mathbf{D}^{\mathrm{sh}}$.

Table~\ref{tab:feature_type_acc} reports the results, which show that the features reconstructed using GB-SAE's shared dictionary produce the best accuracy (Acc@1/Acc@5) over all the other methods. This suggests that GB-SAE extracts richer discriminative information into the representation space associated with its unified dictionary $\mathbf{D}^{\mathrm{sh}}$, as compared to single-model and alignment-based existing methods.

\begin{table}[t]
  \caption{Downstream classification accuracy (Acc@1/Acc@5) on representations extracted by different methods.}
  \label{tab:feature_type_acc}
  \begin{center}
    \begin{small}
      \setlength{\tabcolsep}{4pt}
      \renewcommand{\arraystretch}{0.8}
      \begin{sc}
        \begin{tabular}{lcccc}
          \toprule
          Source & Projected $\uparrow$ & USAE $\uparrow$ & SPARC $\uparrow$ & GB-SAE $\uparrow$ \\
          \midrule
          ViT    & 0.85/0.96 & 0.82/0.96 & 0.84/0.95 & - \\
          SigLIP & 0.75/0.91 & 0.69/0.91 & 0.77/0.93 & - \\
          DinoV2 & 0.69/0.89 & 0.66/0.89 & 0.75/0.91 & - \\
          \midrule
          Mean   & 0.76/0.92 & 0.72/0.92 & 0.78/0.93 & \textbf{0.86/0.97} \\
          \bottomrule
        \end{tabular}
      \end{sc}
    \end{small}
  \end{center}
  \vskip -0.1in
\end{table}

\textbf{Downstream cross-model feature retrieval (full; with training).}
Here we extend the zero-shot retrieval experiment to a full downstream cross-model feature retrieval task; the only difference is that we add a simple retrieval head--a single linear layer only---and train it on representations reconstructed by $\mathbf{D}^{\mathrm{sh}}$ (frozen). The detailed setup and results are provided in Appendix~\ref{app:downstream_retrieval}. The results show that the shared representations learned by GB-SAE substantially outperform all baselines across multiple metrics (instance and category Recalls and mAP), substantiating the strong reusability of the feature representations.

\section{Conclusion}
We introduce the \emph{Gated Bottleneck Sparse Autoencoder} (GB-SAE), a model-agnostic framework for learning a unified semantic dictionary that jointly explains the representations of multiple, heterogeneous neural models. GB-SAE maps heterogeneous representations into a common latent space via a bottleneck and uses a learnable gating mechanism to allocate interpretive capacity between shared semantic concepts and model-specific biases, yielding a unified semantic dictionary alongside private residual components. By instantiating shared semantics in a single reusable latent space—rather than aligning model-specific dictionaries—GB-SAE enables unified interpretability and representation reuse that generalize beyond the architectures seen during training. 
Experiments across three and four pretrained vision models confirm that GB-SAE isolates human-interpretable shared semantics while capturing model-specific structure in structured private residuals. The shared dictionary generalizes to unseen backbones with minimal adaptation, supports zero-shot cross-model retrieval, and yields representations that outperform alignment-based baselines on downstream classification. 

\textbf{Limitations.}
The current evaluation is confined to vision foundation models on ImageNet-scale data. Whether the approach extends to language models, multimodal architectures, or domains with less visual regularity remains an open question. Similarly, our downstream evaluation is limited to linear classification; broader validation across diverse tasks would strengthen the generalization claims.
The pairwise cross-model alignment loss scales as $\mathcal{O}(N^2)$ in the number of models, which adds computational complexity as the ensemble grows. Scaling GB-SAE to large model collections will likely require approximate or hierarchical alignment strategies. 
Finally, the learned decomposition—shared versus private—is soft and governed by the gate regularization strength $\lambda_{\mathrm{gate}}$. The quality of the factorization is sensitive to this hyperparameter, and there is currently no principled criterion for selecting it without empirical tuning.

\begin{ack}
{\bf Funding:} This work was supported in part by computing resources provided by the University of Kentucky Center for Computational Sciences (CCS) and in part by the ECE startup funds provided by Dr. Thomas Tie Luo.

{\bf Competing Interests:} The authors declare no competing interests.
\end{ack}

{
\small
\bibliographystyle{apalike}
\bibliography{refs}
}

% \end{document}
%%%%%%%%%%%%%%%%%%%%%%%%%%%%%%%%%%%%%%%%%%%%%%%%%%%%%%%%%%%%

\newpage
\appendix

\section*{Appendix}

\section{Implementation Details}\label{app:impl}
\textsc{Data.}
We conduct experiments on the standard ImageNet-1K dataset \citep{deng2009imagenet}. Before training, we apply data augmentation to the training set, generating two additional augmented views for each sample. The three views of each sample—the original image and two augmented variants—are then randomly fed into the foundation models to precompute and store their feature representations. GB-SAE is subsequently trained on these static representations. The augmentation pipeline consists of a combination of random resized crop, random horizontal/vertical flips, mild color jittering, and random Gaussian blur. For Configuration I, we use the full ImageNet-1K training and validation splits. For Configurations II and III, we use 10\% of each split, with all 1,000 classes represented. Based on our empirical findings from Configuration I, only marginal differences are observed in the final evaluation metrics between the full ImageNet dataset and its 10\% subset, while the overall trends remain identical.

\textsc{Projection and Encoding.}
Each backbone's output is projected to a common dimensionality of $d' = 768$ using a frozen, randomly initialized linear projector with orthogonal initialization. This design ensures that semantic alignment and disentanglement are learned by GB-SAE itself, rather than being induced by the projection layer. The encoder's latent dimensionality is configured with an eightfold expansion relative to the common dimensionality $d'=768$, resulting in $h=6144$. Top-$K$ sparsification is applied with $K=32$, allowing at most 32 active latent dimensions per input, including both shared and private branches. The exact allocation between $\mathbf{z}^{\mathrm{sh}}$ and $\mathbf{z}^{\mathrm{pr}}$ is governed entirely by the learnable gate.

\textsc{Optimization.}
GB-SAE is trained using AdamW with a learning rate of $3\times10^{-4}$ and batch size 32 for 30 epochs. All backbone networks and projectors are frozen during training. Hyperparameters for the loss weights are selected on the validation set and kept fixed across all experiments. Specifically, we set $\lambda_{\mathrm{sh\text{-}only}}=0.5$, $\lambda_{\mathrm{align}}=1.3$, and $\lambda_{\mathrm{gate}}=0.2$. We fix $\lambda_{\mathrm{mag}}=0.2$, using it as an alignment stabilization term. We conduct additional ablation studies on the three most important loss-weight hyperparameters ($\lambda_{\mathrm{sh\text{-}only}}$, $\lambda_{\mathrm{align}}$, $\lambda_{\mathrm{gate}}$), further details are provided in the Appendix~\ref{app:ablation}.
The results in Table~\ref{tab:recon}(a) and Fig.~\ref{fig:retrieval} are averaged over three random seeds, with the corresponding standard deviations reported in Tables~\ref{tab:reconstruction_std} and ~\ref{tab:retrieval_std}.

\begin{table}[h]
  \caption{\small Standard deviation of reconstruction evaluation on the validation set. Each entry is reported as projected / raw $R^2$ standard deviation.}
  \label{tab:reconstruction_std}
  \begin{center}
    \begin{small}
      \setlength{\tabcolsep}{4pt}
      \renewcommand{\arraystretch}{0.8}
      \begin{sc}
        \begin{tabular}{lcccc}
          \toprule
          Model & $R^2$-total$\downarrow$ & $R^2$-sh$\downarrow$ & $R^2$-pr$\downarrow$ & $R^2$-res$\downarrow$ \\
          \midrule
          ViT    & 0.018/0.018 & 0.023/0.023 & 0.012/0.012 & 0.006/0.006 \\
          SigLIP & 0.019/0.019 & 0.021/0.021 & 0.004/0.004 & 0.009/0.009 \\
          DinoV2 & 0.021/0.021 & 0.031/0.020 & 0.017/0.010 & 0.030/0.019 \\
          \midrule
          Mean   & 0.019/0.019 & 0.025/0.021 & 0.011/0.009 & 0.015/0.011 \\
          \bottomrule
        \end{tabular}
      \end{sc}
    \end{small}
  \end{center}
  \vskip -0.1in
\end{table}

\begin{table}[h]
  \caption{\small Standard deviation of retrieval performance on the validation set. Each entry is reported as Instance / Category retrieval standard deviation.}
  \label{tab:retrieval_std}
  \begin{center}
    \begin{small}
      \setlength{\tabcolsep}{4pt}
      \renewcommand{\arraystretch}{0.8}
      \begin{sc}
        \begin{tabular}{lcccc}
          \toprule
          Metric & R@1$\downarrow$ & R@5$\downarrow$ & R@10$\downarrow$ & R@50$\downarrow$ \\
          \midrule
          PROJ    & 0.000/0.001 & 0.000/0.001 & 0.000/0.001 & 0.000/0.002 \\
          SHARED  & 0.010/0.023 & 0.023/0.038 & 0.029/0.040 & 0.022/0.038 \\
          PRIVATE & 0.000/0.001 & 0.000/0.001 & 0.000/0.002 & 0.001/0.008 \\
          \bottomrule
        \end{tabular}
      \end{sc}
    \end{small}
  \end{center}
  \vskip -0.1in
\end{table}

\textsc{Implementation.}
All experiments are implemented in PyTorch. Training a single GB-SAE model on the Configuration I requires approximately 33 GPU-hours on an NVIDIA H200.

\section{Results under Configurations II and III}\label{app:configs}
\subsection{Results under Configuration II}\label{app:config2}
\begin{table}
  \caption{\small Reconstruction evaluation of GB-SAE with heterogeneous architectures. Each entry is reported as \textit{projected / raw} $R^2$.}
  \label{tab:cnn_recon}
  \begin{center}
    \begin{small}
      \setlength{\tabcolsep}{4pt}
      \renewcommand{\arraystretch}{0.8}
      \begin{sc}
        \begin{tabular}{lcccc}
          \toprule
          Model & $R^2$-total $\uparrow$ & $R^2$-sh $\uparrow$ & $R^2$-pr & $R^2$-res \\
          \midrule
          ViT      & 0.58/0.58 & 0.55/0.55 & 0.08/0.08 & 0.06/0.06 \\
          ConvNeXt & 0.66/0.66 & 0.63/0.63 & 0.06/0.06 & 0.08/0.08 \\
          DinoV2   & 0.64/0.65 & 0.56/0.59 & 0.19/0.21 & 0.18/0.14 \\
          \midrule
          Mean     & 0.63/0.63 & 0.58/0.59 & 0.11/0.12 & 0.11/0.09 \\
          \bottomrule
        \end{tabular}
      \end{sc}
    \end{small}
  \end{center}
  \vskip -0.1in
\end{table}

\begin{figure}
  \centering
  \includegraphics[width=0.7\linewidth]{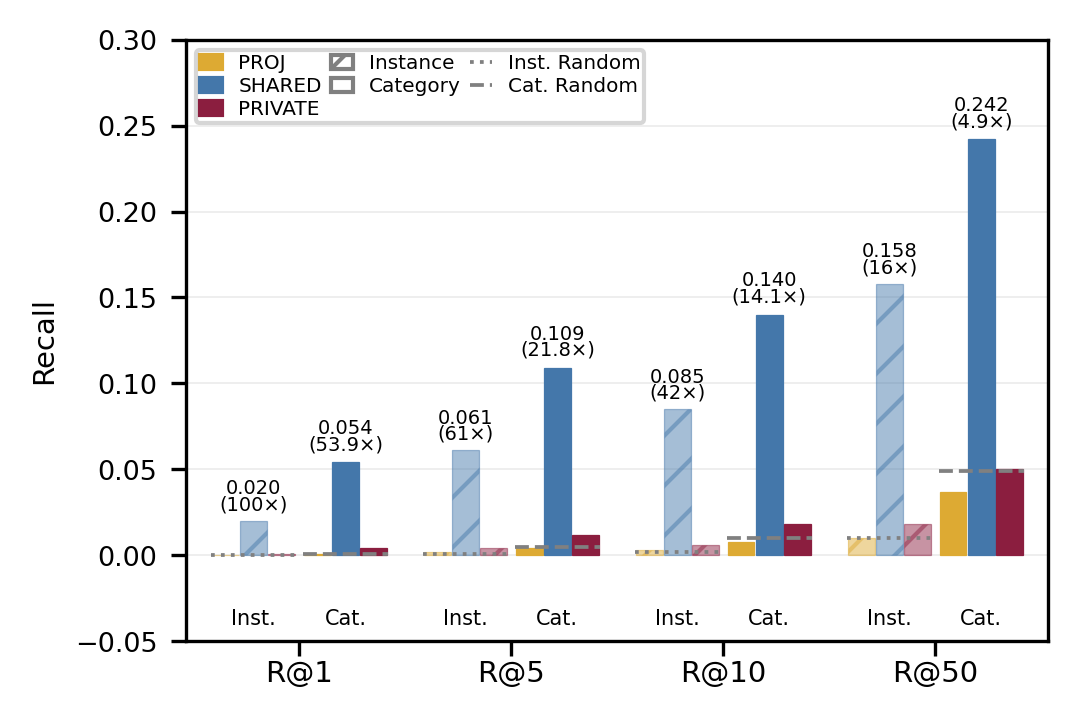}
  \caption{GB-SAE performance on cross-model feature retrieval with heterogeneous model combinations. With the 10\% validation subset, the expected instance-level random Recall@$k$ baseline is approximately 10$\times$ that of Fig.~\ref{fig:retrieval}.}
  \label{fig:cnn_retrieval}
\end{figure}

To further evaluate the performance of GB-SAE on foundation-model combinations with heterogeneous architectures, we conduct experiments on ViT, ConvNeXt, and DINOv2. As shown in Table~\ref{tab:cnn_recon}, GB-SAE effectively accommodates representations from different architectures and achieves strong reconstruction performance. We further perform cross-model feature retrieval under Configuration II to assess the degree of alignment, with the results shown in Figure~\ref{fig:cnn_retrieval}. Overall, GB-SAE remains effective under heterogeneous model configurations, demonstrating its ability to factorize shared concepts from model-specific biases and extract a unified cross-model semantic dictionary.

\subsection{Results under Configuration III}\label{app:config3}
\begin{table}
  \caption{\small Reconstruction evaluation of GB-SAE under Configuration III. Each entry is reported as \textit{projected / raw} $R^2$.}
  \label{tab:config3_recon}
  \begin{center}
    \begin{small}
      \setlength{\tabcolsep}{4pt}
      \renewcommand{\arraystretch}{0.8}
      \begin{sc}
        \begin{tabular}{lcccc}
          \toprule
          Model & $R^2$-total $\uparrow$ & $R^2$-sh $\uparrow$ & $R^2$-pr & $R^2$-res \\
          \midrule
          ViT      & 0.57/0.57 & 0.52/0.52 & 0.11/0.11 & 0.11/0.11 \\
          SigLIP   & 0.58/0.58 & 0.51/0.51 & 0.21/0.21 & 0.15/0.15 \\
          DINOv2   & 0.65/0.66 & 0.52/0.58 & 0.21/0.24 & 0.27/0.20 \\
          ConvNeXt & 0.65/0.65 & 0.60/0.60 & 0.06/0.06 & 0.13/0.13 \\
          \midrule
          Mean     & 0.62/0.62 & 0.54/0.55 & 0.15/0.16 & 0.16/0.14 \\
          \bottomrule
        \end{tabular}
      \end{sc}
    \end{small}
  \end{center}
  \vskip -0.1in
\end{table}

\begin{figure}
  \centering
  \includegraphics[width=0.7\linewidth]{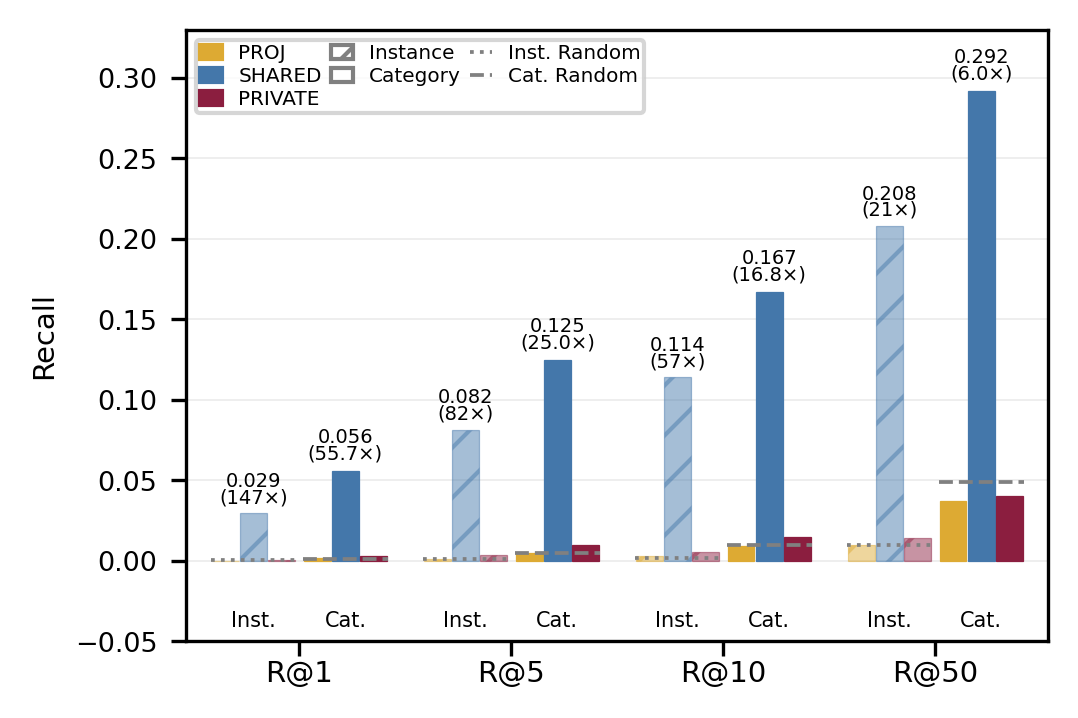}
  \caption{Cross-model feature retrieval performance under Configuration III. We report Recall@$k$ at the instance and category levels. With the 10\% validation subset, the instance-level random baseline is approximately 10$\times$ that of Fig.~\ref{fig:retrieval}.}
  \label{fig:config3_retrieval}
\end{figure}

To assess how GB-SAE behaves as the ensemble expands from three to four heterogeneous backbones, Configuration III combines ViT, SigLIP, DINOv2, and ConvNeXt. Compared with Configuration II, mean $R^2$-total and $R^2$-sh decrease modestly (0.63/0.63 to 0.62/0.62 and 0.58/0.59 to 0.54/0.55, respectively), while mean $R^2$-pr and $R^2$-res increase (0.11/0.12 to 0.15/0.16 and 0.11/0.09 to 0.16/0.14; Tables~\ref{tab:cnn_recon} and \ref{tab:config3_recon}). This trend is expected: adding another model makes it harder for the shared dictionary to capture structure common to every backbone, while the private branch accounts for a larger share of model-specific or less widely shared variation. Despite this shift, Configuration III retains substantial reconstruction quality, with mean $R^2$-total of 0.62/0.62 and mean $R^2$-sh of 0.54/0.55 (projected/raw). Instance Recall@1/5/10/50 is 0.03/0.08/0.11/0.21, while Category Recall@1/5/10/50 is 0.06/0.13/0.17/0.29 (Fig.~\ref{fig:config3_retrieval}), indicating that the shared space remains useful in the four-model setting.
% We also examine the computational cost of adding backbone models to GB-SAE. As discussed in Section~\ref{sec:training_objective}, the cross-model alignment objective introduces a quadratic term in the number of models, which could become the dominant source of additional cost as the ensemble grows. In practice, however, our experiments show that this quadratic term does not incur significant computational overhead at the model scales tested. Formally, for $N$ models, batch size $B$, $T$ tokens, latent width $h$, and projected dimension $d'$, the alignment cost is $\mathcal{O}(N^2 B T h)$, whereas SAE encoding and decoding scale as $\mathcal{O}(N B T d' h)$. Although the number of model pairs grows quadratically, each pair requires only lightweight operations such as normalization and min/max, while the encoder and decoders involve more expensive dense matrix multiplications. Thus, within a reasonable range of model counts, alignment is not the dominant computational bottleneck. In our experiments, the training time for Configuration III was almost unchanged relative to Configuration II. Nevertheless, the quadratic term may become significant for very large model collections (which would be atypical). In that setting, randomly sampling a subset of model pairs per mini-batch or aligning each model to an anchor can reduce the complexity to approximately $\mathcal{O}(N)$ or $\mathcal{O}(N\log N)$, respectively, enabling large-scale GB-SAE training.

\section{Qualitative Visualization}\label{app:vis}
\subsection{Algorithm}
\begin{algorithm}[h]
  \caption{GB-SAE Concept Visualization}
  \label{alg:gbsae_vis}
  \begin{algorithmic}
    \STATE {\bfseries Input:} Trained GB-SAE, threshold $\tau_{\mathrm{sh}} = 0.5$, dataset $\mathcal{X}$
    \STATE {\bfseries Output:} Token-level heatmaps of concept atoms
    \STATE Load the trained GB-SAE model
    \STATE $\mathbf{g} \gets \mathrm{GetGate}(\text{GB-SAE})$
    \FOR{each concept latent index $h$}
        \IF{$\mathbf{g}_h > \tau_{\mathrm{sh}}$}
            \STATE Assign $h \rightarrow \mathbf{h}_{\mathrm{sh}}$
        \ELSE
            \STATE Assign $h \rightarrow \mathbf{h}_{\mathrm{pr}}$
        \ENDIF
    \ENDFOR
    \FOR{each input sample $x \in \mathcal{X}$}
        \FOR{each $h \in \mathbf{h}_{\mathrm{sh}} / \mathbf{h}_{\mathrm{pr}}$} 
            \STATE $(\bar{\mathbf{z}}^{\mathrm{sh}}_{x,h}, \mathbf{z}^{\mathrm{pr}}_{x,h}) \gets \mathrm{GB\text{-}SAE}(x)$
            \STATE Generate TopN token-level heatmap for each concept atom $h$
        \ENDFOR
    \ENDFOR
    \STATE {\bfseries Return:} All heatmaps
  \end{algorithmic}
\end{algorithm}

The Algorithm \ref{alg:gbsae_vis} illustrates the procedure for visualizing concepts. First, we use 0.5 as the gating threshold to select the shared/private concepts to be visualized. We then obtain the mean activation values $\bar{\mathbf{z}}^{\mathrm{sh}}_{x,h}$ of shared concepts on each sample, and the model-specific activation values $\mathbf{z}^{\mathrm{pr}}_{x,h}$ of private concepts on each sample. Finally, for each concept, we select the top-$N$ samples with the largest overall activation values, namely those that most strongly activate the concept, and use the corresponding maximum activations to generate their token-level heatmaps.

\subsection{Visualization of Private Concepts}
\begin{figure}[h]
  \centering

  \begin{minipage}{0.47\linewidth}
    \centering
    \includegraphics[width=\linewidth]{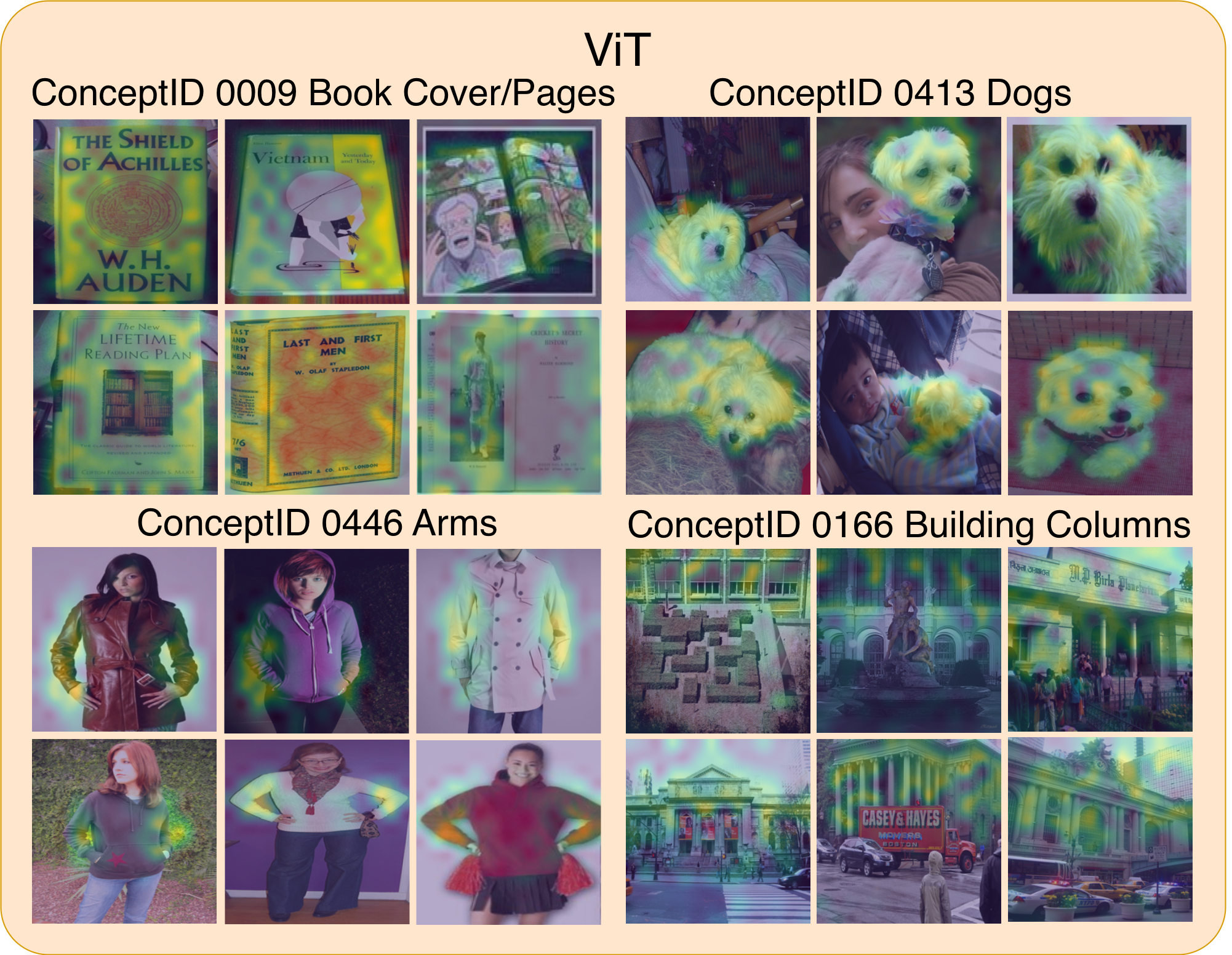}
  \end{minipage}
  \hfill
  \begin{minipage}{0.47\linewidth}
    \centering
    \includegraphics[width=\linewidth]{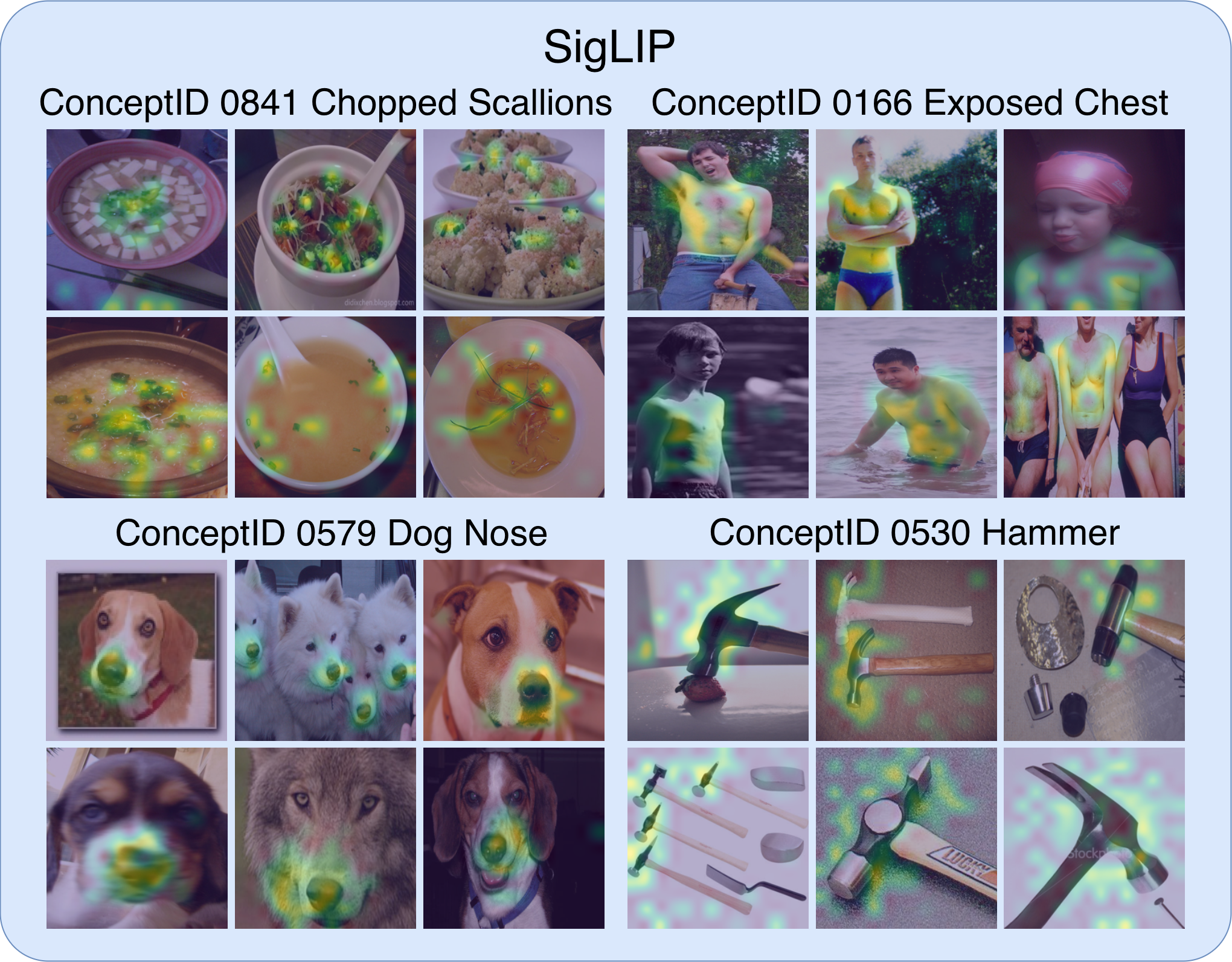}
  \end{minipage}
  \hfill
  \begin{minipage}{0.47\linewidth}
    \centering
    \includegraphics[width=\linewidth]{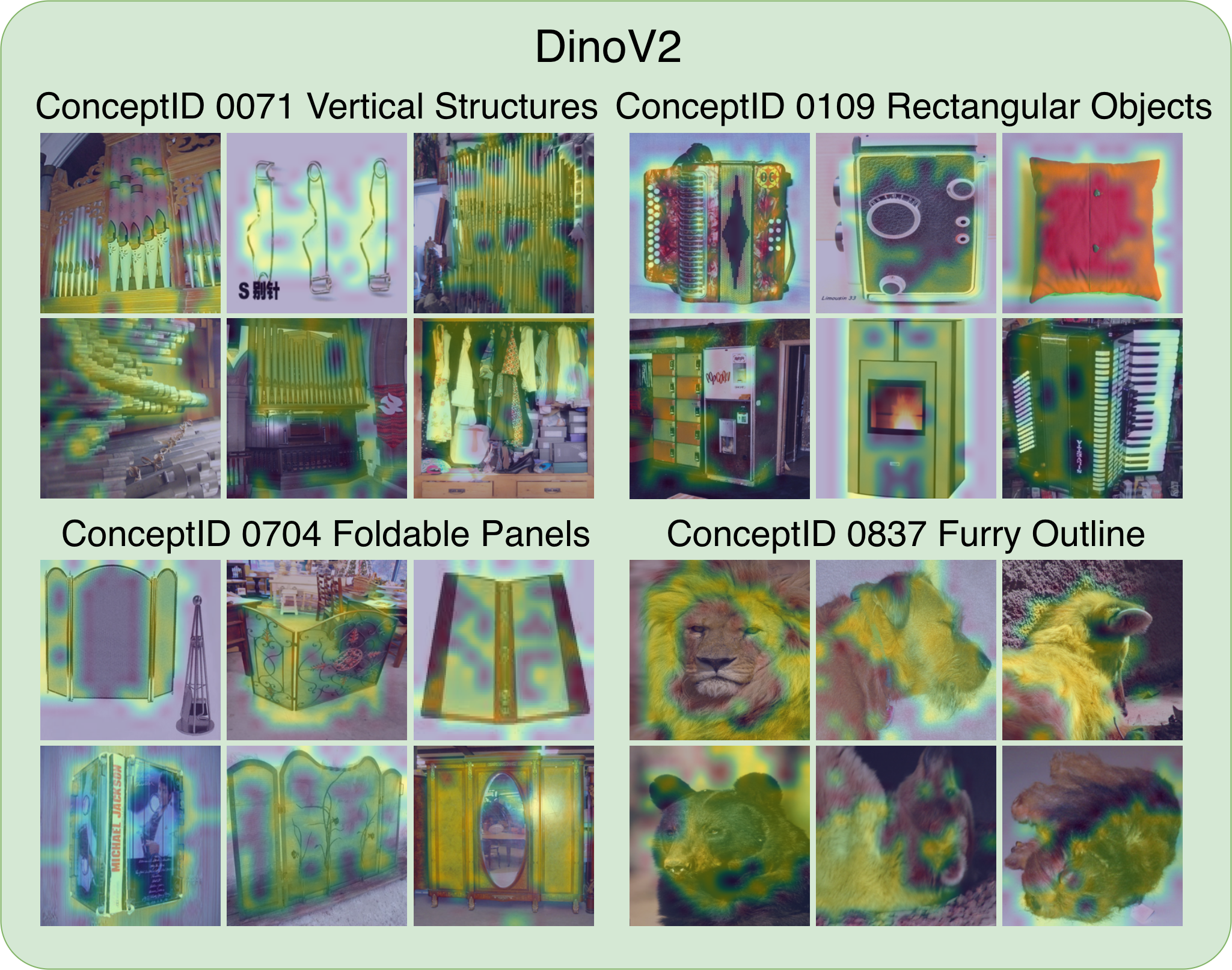}
  \end{minipage}

  \caption{Visualization of model-specific private concepts. Unlike shared atoms, private concepts reflect architecture- and training-dependent inductive biases, exhibiting stronger locality and model-specific activation patterns.}
  \label{fig:visual-private}
\end{figure}

Fig.~\ref{fig:visual-private} presents the visualization results of private concepts extracted from different backbones. For ViT, private concepts mainly capture class-discriminative textures and shape fragments, reflecting supervised classification biases. SigLIP’s private atoms align with semantically coherent visual prototypes tied to language, highlighting the effect of vision–language pretraining. In contrast, DINOv2’s private concepts emphasize geometric structures and spatial patterns, consistent with self-supervised objectives that favor geometric invariances over semantic labels.

\section{Ablation Study}\label{app:ablation}
% \subsection{Loss-weight ablation}
\begin{figure}[h]
  \centering

  \begin{minipage}{1.0\linewidth}
    \centering
    \includegraphics[width=\linewidth]{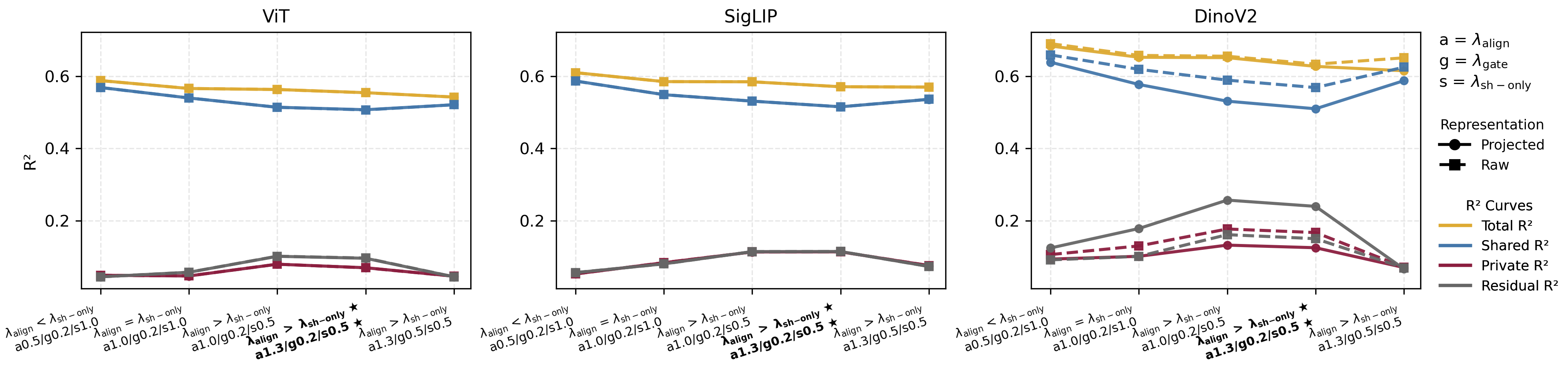}
    \centerline{(a) Reconstruction results}
  \end{minipage}
  \vspace{0.8em}
  \begin{minipage}{1.0\linewidth}
    \centering
    \includegraphics[width=\linewidth]{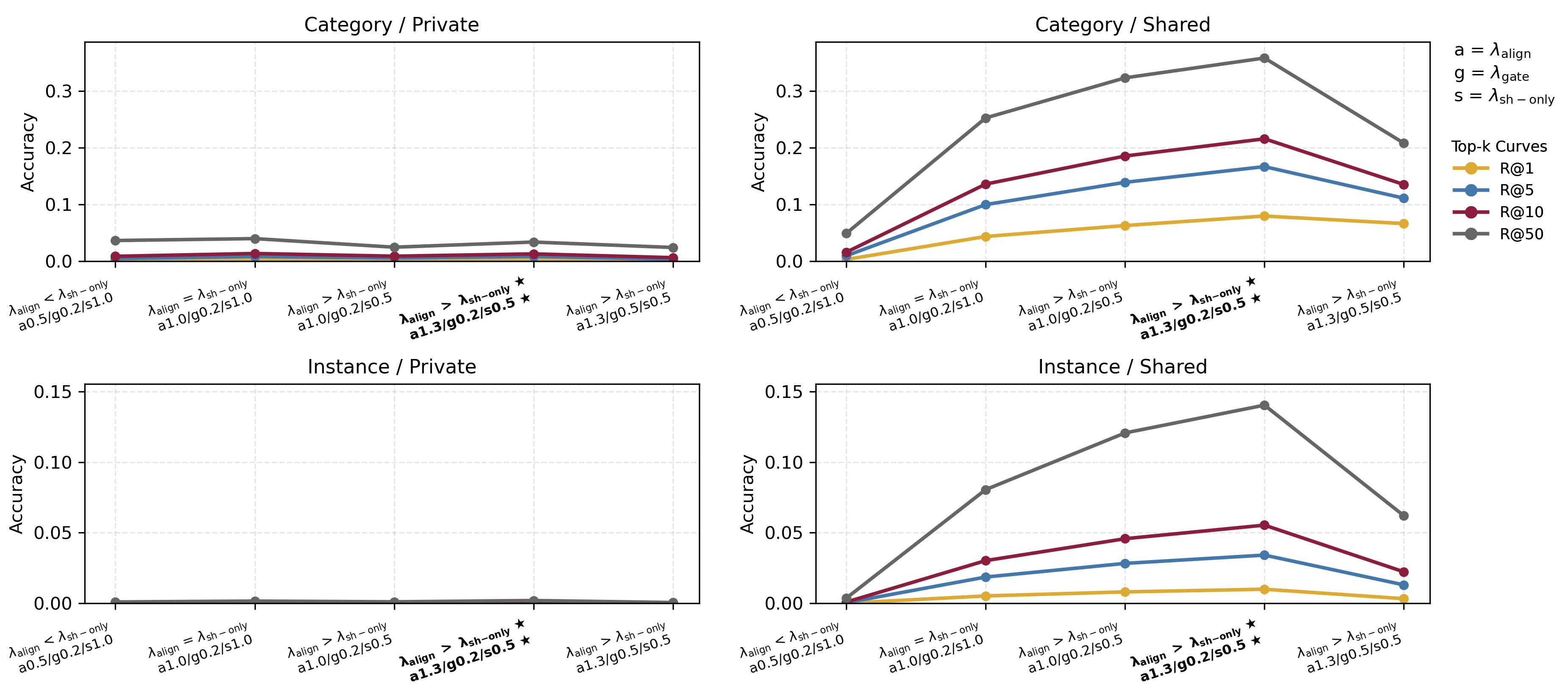}
    \centerline{(b) Alignment results}
  \end{minipage}
  \caption{GB-SAE loss-weight ablation study. Figure (a) shows the reconstruction results under different configurations, while Figure (b) reports the corresponding alignment results.}
  \label{fig:ablation}
\end{figure}

The proposed method primarily achieves semantic disentanglement through the combination of a bottleneck-style architecture and three loss functions. Therefore, the relative weights of these loss terms are particularly important for the disentanglement performance of GB-SAE. We consider several weight configurations to examine the importance and rationale of each component: (i) with the $\lambda_{\mathrm{gate}}$ fixed, we set the $\lambda_{\mathrm{align}}$ to be larger than, equal to, or smaller than the $\lambda_{\mathrm{sh\text{-}only}}$; and (ii) with the $\lambda_{\mathrm{align}}$ and$\lambda_{\mathrm{sh\text{-}only}}$ fixed, we test both smaller and larger values of the $\lambda_{\mathrm{gate}}$. We do not evaluate the setting $\lambda_{\mathrm{gate}}=0$, which removes the regularization encouraging binary gate values while retaining the gating mechnaism itself.

The ablation results are shown in Figure~\ref{fig:ablation}. We observe that the $\lambda_{\mathrm{sh\text{-}only}}$ and $\lambda_{\mathrm{align}}$ consistently exhibit a trade-off. A larger $\lambda_{\mathrm{sh\text{-}only}}$ preserves the reconstruction ability of the shared branch, but substantially weakens cross-model retrieval performance; conversely, a larger $\lambda_{\mathrm{align}}$ also degrades reconstruction quality. In addition, the $\lambda_{\mathrm{gate}}$ acts as a structural optimization pressure: when it is too large, the gates rapidly converge to closed states, severely impairing alignment. These results suggest that the three loss terms play complementary roles and must be carefully balanced to achieve effective shared/private decomposition and cross-model alignment.
% \subsection{Toward More General Representation Learning Scenarios}
% \begin{figure}[h]
%   \centering
%   \includegraphics[width=0.65\linewidth]{LargeK.png}
%   \caption{Performance of GB-SAE under progressively relaxed sparsity constraints. As sparsity is reduced, the shared dictionary continues to absorb additional semantic structure, indicating that GB-SAE scales beyond sparse concept discovery toward more general representation unification.}
%   \label{largeK}
% \end{figure}

% The preceding results establish that the shared dictionary $\mathbf{D}^{\mathrm{sh}}$ learned by GB-SAE captures transferable and model-agnostic semantics under a sparse, interpretable regime. Beyond concept discovery, however, a natural question is: {\em can the same framework support more general representation learning objectives?}

% To this end, we relax the sparsity constraint, allowing more latent units to participate in reconstruction. This shifts GB-SAE from a strictly sparse concept extractor toward a multi-model representation alignment and reconstruction framework, where a larger dictionary can encode increasingly finer-grained semantics. We progressively reduce sparsity and retrain GB-SAE under identical settings (Fig.~\ref{largeK}).

% The results show that $\mathbf{D}^{\mathrm{sh}}$ continues to absorb additional semantic detail as sparsity is relaxed, rather than saturating. This suggests that GB-SAE scales beyond interpretability and can serve as a general framework for unifying heterogeneous representations into a rich, reusable semantic substrate.

\section{Downstream Task: Retrieval}\label{app:downstream_retrieval}
We evaluate cross-model retrieval using four types of frozen image representations and train retrieval heads for each method. PROJECTED uses backbone activations after the frozen linear projectors and token mean pooling. USAE and SPARC use the mean-pooled representations reconstructed by their respective SAEs. The retrieval heads for PROJECTED, USAE, and SPARC are trained on the full original or reconstructed representations. In contrast, GB-SAE uses only the shared component, obtained by encoding each model’s projected token activations as \(\mathbf{z}^{\mathrm{sh}}_i\), decoding them with the shared dictionary \(\mathbf{D}^{\mathrm{sh}}\), mean-pooling the reconstructions, and averaging the reconstructions to obtain the fused shared representation.

The retrieval heads are trained to align representations of the same image across models while encouraging representations of different images from the same class to be closer in the embedding space. As shown in Table~\ref{tab:downstream_retrieval}, GB-SAE achieves the best performance across all reported metrics. 

\begin{table}[h]
  \caption{\small Downstream cross-model retrieval performance using four frozen image representations with trainable linear retrieval heads; best result per metric is bold. Instance R@1 measures whether the same image is ranked first. Category R@K removes the paired same image and measures whether another image of the same class appears in the top $K$. For each query, Category AP averages the precision at every rank containing a same-class image; mAP is the mean AP over all valid queries, so it evaluates the quality of the full ranking rather than only the first hit.}
  \label{tab:downstream_retrieval}
  \begin{center}
    \begin{small}
      \setlength{\tabcolsep}{4pt}
      \renewcommand{\arraystretch}{0.9}
      \begin{sc}
        \begin{tabular}{lcccc}
          \toprule
          Metric & Projected & USAE & SPARC & GB-SAE \\
          \midrule
          Instance R@1 & 0.96 & 0.91 & 0.47 & \textbf{0.99} \\
          Category R@1 & 0.48 & 0.44 & 0.52 & \textbf{0.57} \\
          Category R@5 & 0.75 & 0.72 & 0.74 & \textbf{0.81} \\
          Category mAP & 0.39 & 0.37 & 0.44 & \textbf{0.48} \\
          \bottomrule
        \end{tabular}
      \end{sc}
    \end{small}
  \end{center}
\end{table}

%%%%%%%%%%%%%%%%%%%%%%%%%%%%%%%%%%%%%%%%%%%%%%%%%%%%%%%%%%%%

\newpage
\section*{NeurIPS Paper Checklist}

\begin{enumerate}

\item {\bf Claims}
    \item[] Question: Do the main claims made in the abstract and introduction accurately reflect the paper's contributions and scope?
    \item[] Answer: \answerYes{} % Replace by \answerYes{}, \answerNo{}, or \answerNA{}.
    \item[] Justification: The abstract and introduction accurately summarize the paper's contributions, scope, and limitations, and the claims are supported by the theoretical analysis and experimental results presented in the paper.

\item {\bf Limitations}
    \item[] Question: Does the paper discuss the limitations of the work performed by the authors?
    \item[] Answer: \answerYes{} % Replace by \answerYes{}, \answerNo{}, or \answerNA{}.
    \item[] Justification: The limitations of the proposed method are discussed in the Conclusion section.

\item {\bf Theory assumptions and proofs}
    \item[] Question: For each theoretical result, does the paper provide the full set of assumptions and a complete (and correct) proof?
    \item[] Answer: \answerNA{} % Replace by \answerYes{}, \answerNo{}, or \answerNA{}.
    \item[] Justification: The paper does not include theoretical results or formal proofs.

    \item {\bf Experimental result reproducibility}
    \item[] Question: Does the paper fully disclose all the information needed to reproduce the main experimental results of the paper to the extent that it affects the main claims and/or conclusions of the paper (regardless of whether the code and data are provided or not)?
    \item[] Answer: \answerYes{} % Replace by \answerYes{}, \answerNo{}, or \answerNA{}.
    \item[] Justification: The paper provides sufficient experimental details, including datasets, evaluation metrics, implementation settings, and hyperparameters, to reproduce the main results.

\item {\bf Open access to data and code}
    \item[] Question: Does the paper provide open access to the data and code, with sufficient instructions to faithfully reproduce the main experimental results, as described in supplemental material?
    \item[] Answer: \answerYes{} % Replace by \answerYes{}, \answerNo{}, or \answerNA{}.
    \item[] Justification: The code has been uploaded to the Supplementary Material.

\item {\bf Experimental setting/details}
    \item[] Question: Does the paper specify all the training and test details (e.g., data splits, hyperparameters, how they were chosen, type of optimizer) necessary to understand the results?
    \item[] Answer: \answerYes{} % Replace by \answerYes{}, \answerNo{}, or \answerNA{}.
    \item[] Justification: The paper specifies the experimental settings, including datasets, training details, hyperparameters, and evaluation protocols necessary to understand and reproduce the results.

\item {\bf Experiment statistical significance}
    \item[] Question: Does the paper report error bars suitably and correctly defined or other appropriate information about the statistical significance of the experiments?
    \item[] Answer: \answerYes{} % Replace by \answerYes{}, \answerNo{}, or \answerNA{}.
    \item[] Justification: We conduct experiments with three random seeds and report the average results in the main paper, with the corresponding standard deviations provided in Appendix A.

\item {\bf Experiments compute resources}
    \item[] Question: For each experiment, does the paper provide sufficient information on the computer resources (type of compute workers, memory, time of execution) needed to reproduce the experiments?
    \item[] Answer: \answerYes{} % Replace by \answerYes{}, \answerNo{}, or \answerNA{}.
    \item[] Justification: The paper reports the compute resources used for the experiments in Appendix A.
    
\item {\bf Code of ethics}
    \item[] Question: Does the research conducted in the paper conform, in every respect, with the NeurIPS Code of Ethics \url{https://neurips.cc/public/EthicsGuidelines}?
    \item[] Answer: \answerYes{} % Replace by \answerYes{}, \answerNo{}, or \answerNA{}.
    \item[] Justification: The research follows the NeurIPS Code of Ethics and does not involve human subjects, private data, or other ethical concerns.

\item {\bf Broader impacts}
    \item[] Question: Does the paper discuss both potential positive societal impacts and negative societal impacts of the work performed?
    \item[] Answer: \answerNo{} % Replace by \answerYes{}, \answerNo{}, or \answerNA{}.
    \item[] Justification: This work primarily focuses on the technical aspects of the proposed method and does not include a dedicated discussion of broader societal impacts.
    
\item {\bf Safeguards}
    \item[] Question: Does the paper describe safeguards that have been put in place for responsible release of data or models that have a high risk for misuse (e.g., pre-trained language models, image generators, or scraped datasets)?
    \item[] Answer: \answerNA{} % Replace by \answerYes{}, \answerNo{}, or \answerNA{}.
    \item[] Justification: The paper does not release data or models that pose a high risk of misuse.

\item {\bf Licenses for existing assets}
    \item[] Question: Are the creators or original owners of assets (e.g., code, data, models), used in the paper, properly credited and are the license and terms of use explicitly mentioned and properly respected?
    \item[] Answer: \answerYes{} % Replace by \answerYes{}, \answerNo{}, or \answerNA{}.
    \item[] Justification: All datasets, models, and code assets used in this work are properly credited and used in accordance with their respective licenses and terms of use.

\item {\bf New assets}
    \item[] Question: Are new assets introduced in the paper well documented and is the documentation provided alongside the assets?
    \item[] Answer: \answerNA{} % Replace by \answerYes{}, \answerNo{}, or \answerNA{}.
    \item[] Justification: The paper does not release new datasets, models, or other assets.

\item {\bf Crowdsourcing and research with human subjects}
    \item[] Question: For crowdsourcing experiments and research with human subjects, does the paper include the full text of instructions given to participants and screenshots, if applicable, as well as details about compensation (if any)? 
    \item[] Answer: \answerNA{} % Replace by \answerYes{}, \answerNo{}, or \answerNA{}.
    \item[] Justification: The paper does not involve crowdsourcing or research with human subjects.

\item {\bf Institutional review board (IRB) approvals or equivalent for research with human subjects}
    \item[] Question: Does the paper describe potential risks incurred by study participants, whether such risks were disclosed to the subjects, and whether Institutional Review Board (IRB) approvals (or an equivalent approval/review based on the requirements of your country or institution) were obtained?
    \item[] Answer: \answerNA{} % Replace by \answerYes{}, \answerNo{}, or \answerNA{}.
    \item[] Justification: The paper does not involve research with human subjects or crowdsourcing experiments.

\item {\bf Declaration of LLM usage}
    \item[] Question: Does the paper describe the usage of LLMs if it is an important, original, or non-standard component of the core methods in this research? Note that if the LLM is used only for writing, editing, or formatting purposes and does \emph{not} impact the core methodology, scientific rigor, or originality of the research, declaration is not required.
    %this research? 
    \item[] Answer: \answerNA{} % Replace by \answerYes{}, \answerNo{}, or \answerNA{}.
    \item[] Justification: The core methodology of this research does not rely on LLMs as an important or non-standard component.

\end{enumerate}

\end{document}